\documentclass{article}

\usepackage{arxiv}

\usepackage[utf8]{inputenc} 
\usepackage[T1]{fontenc}    
\usepackage{hyperref}       
\usepackage{url}            
\usepackage{booktabs}       
\usepackage{amsfonts}       
\usepackage{nicefrac}       
\usepackage{microtype}      
\usepackage{lipsum}

\usepackage{times}
\usepackage{epsfig}
\usepackage{graphicx}
\usepackage{amsmath}
\usepackage{amssymb}

\usepackage{array}
\usepackage{multirow}
\usepackage{amsmath}
\usepackage{subfigure}
\usepackage{nicefrac}
\usepackage{caption}
\usepackage{color,soul}
\usepackage{eqnarray,amsmath}
\usepackage{algorithm}
\usepackage{algpseudocode}

\usepackage{framed} 
\usepackage{color,soul}
\usepackage{rotating}
\usepackage[normalem]{ulem}
\usepackage{arydshln}
\usepackage{url}
\usepackage{hyperref} 
\usepackage{comment}

\newcommand{\alert}[1]{\textcolor{red}{#1}}

\newcommand*{\Comb}[2]{{}^{#1}C_{#2}} 

\usepackage{mathtools, cuted}
\newcounter{myEqnCNT}
\newcommand{\myEqnCnt}[2]{\refstepcounter{myEqnCNT}\label{#1}\arabic{myEqnCNT}}

\title{Posture and Sequence Recognition for {\em Bharatanatyam} Dance Performances using Machine Learning Approach}

\author{
  Tanwi~Mallick, Partha Pratim Das, Arun Kumar Majumdar \\
  Department of Computer Science and Engineering\\
  Indian Institute of Technology, Kharagpur,
  India, 721302 \\
  \texttt{tanwimallick@gmail.com, ppd@cse.iitkgp.ac.in, akmj@cse.iitkgp.ac.in} \\
}

\begin{document}
\maketitle

\begin{abstract}
Understanding the underlying semantics of performing arts like dance is a challenging task. Dance is multimedia in nature and spans over time as well as space. Capturing and analyzing the multimedia content of dance is useful for preservation of cultural heritage, to build video recommendation systems, to assist learners use tutoring systems etc. To develop an application for dance, three aspects of dance analysis need to be addressed: 1) Segmentation of the dance video to find the representative action elements, 2) Matching or recognition of the detected action elements, and 3) Recognition of the dance sequences formed by combining a number of action elements under certain rules. This paper attempts to solve three fundamental problems of dance analysis for understanding the underlying semantics of dance forms. Our focus is on an {\em Indian Classical Dance} (ICD) form known as {\em Bharatanatyam}. As dance is driven by music, we use the musical as well as motion information for key posture extraction. Next, we recognize the key postures using machine learning as well as deep learning techniques. Finally, the dance sequence is recognized using Hidden Markov Model (HMM). We capture the multi-modal data of {\em Bharatanatyam} dance using Kinect and build an annotated data set for research in ICD.
\end{abstract}


\section{Introduction}
\label{sec:introduction}

Dance is a specific form of human activity. Hence dance analysis is a special kind of human activity analysis~\cite{xu2017hierarchical, wang2017context, guo2016tensor}. While a human activity can be something like making tea, shaking hands, cleaning windows, doing shopping etc., mere combination of a few movements does not make a dance. Dance is a language of communication  driven by music and follows a set of rules that depends on its specific form. In dance, a series of unique and elementary activity units (action elements) combines with music to embody an idea or to express emotions. With the rapid advances in technologies, we are now able to capture, analyze, and interpret the complex movements of dance. 

Sparse yet varied research has been carried out on dance analysis to develop applications such as video recommendation~\cite{han2017dancelets} based on dance styles and performances, dance tutoring system~\cite{alexiadis2014quaternionic}, dance video annotation~\cite{mallik2011nrityakosha} for heritage preservation, and music-driven dance synthesis~\cite{ofli2012learn2dance}. In these applications, the authors attempt to solve a combination of the three fundamental problems of dance analysis: 1) Segmentation of the dance video in terms of the representative action elements, 2) Matching or recognition of the detected action elements, and 3) Recognition of the dance sequences formed by combination of a number of action elements under certain rules. Han et al.~\cite{han2017dancelets} build a dance video recommendation system. They find the most discriminating action elements between the different dance forms using Normalized Cut Clustering and Linear Discriminant Analysis. Further, random forest is used for fast matching of action elements to generate recommendations. Alexiadis~\cite{alexiadis2014quaternionic} proposes a system for evaluating dance performances of novice against a expert dancer. The temporal synchronization of the motion data of two dancers is achieved using {\em Dynamic Time Warping} (DTW). Finally, they use a set of quaternionic correlation-based measures (scores) for matching and comparing the action elements of two dancers. Mallik et al.~\cite{mallik2011nrityakosha} use {\em Multimedia Web Ontology Language} (MOWL) to represent the domain knowledge of {\em Indian Classical Dance} (ICD). They construct the ontology with a labeled set of training data. Given a new dance video, the system recognizes the spatio-temporal action elements and music primitives, and label them using the ontological concepts. They use clustering to find the primitive action elements. Ofli et al.~\cite{ofli2012learn2dance} propose a framework for music-driven dance analysis to synthesize choreography. Dance has been intimately associated with music for generations. Given a dance video, the proposed model learns the mapping between the music primitives and the dance action elements. The mapping thus learned, is later used for synthesizing new choreography. The authors segment the music signal into music primitives and consider the corresponding motion trajectory as a single action element. To find the mapping they extract the features of the music primitives as well as the action elements and, based on a {\em Hidden Markov Model} (HMM), establish the final mapping for choreography synthesis. 

Given the state of the art, our paper attempts to  investigate the three fundamental problems related to modeling and capture of the underlying semantics of dance. While the existing research attempts to solve a subset of these problems to support specific applications such as video recommendation~\cite{han2017dancelets} based on dance styles and performances, dance tutoring system~\cite{alexiadis2014quaternionic}, dance video annotation~\cite{mallik2011nrityakosha} for heritage preservation, or music-driven dance synthesis~\cite{ofli2012learn2dance}; here we try to solve all the three problems in unison to get a good grasp on the formation of a dance sequence and its underlying semantics. This is particularly important since ICD involves complex combinations of music, movements and gestures that needs the solution to all the three problems for a comprehensive analysis. 

We develop our system based on a form of ICD known as {\em Bharatanatyam}. Analysis of complete {\em Bharatanatyam} performances with all its generic features and idiosyncrasies is an extremely complex task. Hence, to keep the complexity of the problem manageable, we work only with {\em Adavu}s of {\em Bharatanatyam} in this paper. An {\em Adavu} is a basic unit of {\em Bharatanatyam} performance comprising well-defined sets of postures, gestures, movements and their transitions. It is traditionally used to train the dancers.

We address the fundamental problems mentioned above in multiple steps. As a dance video comprises of long sequence of frames (many of which are quite similar), the first challenge is to find a set of most representative and discriminating {\em action elements} that can be used to construct a best summary of the associated frame sequence. Hence our first task is to segment the dance videos and to extract {\em key action elements}. Such segmentation, however, would usually be subjective and may vary depending on the style of dance, or the dancer, or both. Current literature on posture and gesture recognition research~\cite{mohanty2016nrityabodha, samanta2012indian, guo2006dance} often does not consider this issue of segmentation and assumes that pre-segmented sequences are available for analysis. In contrast, Kahol et al. propose a methodology in~\cite{kahol2004automated} for automated segmentation of action elements from dance sequences. They use velocity, acceleration and mass of the various segments of the whole body to represent an activity. Using the measures of activities the dance motion is parsed into discrete units. The authors claim 93.3\% accuracy for \ the detected gesture boundaries. 
Sharma~\cite{apratimrecog} and Han et al.~\cite{han2017dancelets} extract the action elements by clustering. In~\cite{shiratori2003rhythmic, shiratori2004detecting}, the authors propose methods to use musical information for motion structure analysis and segmentation. They detect the onset from the music signal, track the beats, and refine the motion based gesture segmentation using the musical beats. 

The next challenge is to recognize the extracted action elements. In some dance forms the action elements are small signature movements or gesture such as {\em plie} or {\em releve} in {\em Ballet}~\cite{campbell1995recognition}. 
However, in a {\em Bharatanatyam Adavu} the {\em key action elements} of a dance sequence are momentarily stationary postures. We refer to them as {\em Key Posture}s (KPs). Contrary to intuition, {\em Key Posture} recognition is rather non-trivial due to several factors. First, the input has high dimensionality and huge variability in acceptable postures. Second, the dancers often wear long dresses hiding major body parts. Finally, many postures are quite complex and few body parts can get occluded in the frontal view. 
In~\cite{guo2006dance} and~\cite{peng2008binocular} orthogonal stereo cameras are used to capture the postures of a contemporary dance. Two views have been represented and used as features for recognition. {\em Relevance Vector Machine} (RVM) and {\em Support Vector Machine} (SVM) are deployed for posture recognition. Recognition of ICD postures has been explored by~\cite{mohanty2016nrityabodha} using Kinect data as well as RGB images. The authors show that deep learning using {\em Convolutional Neural Network} (CNN) performs better than the conventional approach of using {\em Histogram of Oriented Gradient} (HOG) features with SVM classifier. They use two data sets (one recorded at the laboratory and the other collected from YouTube) of 12 and 14 postures with 720 and 1008 training images, and 144 and 252 test images respectively. The recognition accuracy using HOG + SVM on background-eliminated binary images is reported as 86.11\% and 88.89\% respectively on the two sets. In contrast, CNN achieves 98.60\% and 98.40\% accuracy on the same data sets. While little work exist on the posture recognition in ICD, there is sizeable literature on pose recognition in general including~\cite{andriluka2009pictorial, ning2008discriminative, johnson2011learning, tian2012exploring}, and~\cite{dantone2013human}. However, most of these are not directly applicable in our context.

The final challenge is to recognize the dance sequence. A dance sequence is like a sentence comprising a set of action elements (or KPs in our case) in a specific order. In {\em Bharatanatyam} each {\em Adavu} represents a different dance sequence. The rules for composing an {\em Adavu} using KPs are defined in {\em Bharatanatyam}. The challenge, here, is to capture the rule set in the classifier and recognize an unknown {\em Adavu} using the trained classifier. Sharma in~\cite{apratimrecog} presents a method to recognize 12 {\em Adavu}s of {\em Bharatanatyam} using Kinect RGB-D data. Every {\em Adavu} is represented as a sequence of postures and characterized by the postures present in the {\em Adavu} and their order. A posture is represented by skeleton angles. The data set of skeleton angles is clustered to create a dictionary of postures using {\em Gaussian Mixture Model} (GMM). The postures are described in terms of histogram of postures and then the histogram is learnt using an SVM classifier. But this does not take into account the temporal information or the audio, and just relies on the occurrence of a posture. However, the order of occurrence of any posture is important for the representation of any {\em Adavu}. Hence, ~\cite{apratimrecog} uses HMM to preserve the temporal information of an action sequence. The method achieves 80.55\% cross validation accuracy for {\em Adavu} recognition. This work does not consider the inherent structural information of ICD as the authors do not consider the music. They extract the key postures by clustering which lead to huge number of posture classes. However, this is the only work on classification of ICD sequences. Ofli et al.~\cite{ofli2012learn2dance} capture the intrinsic dependencies of dance postures of Turkish folk dance using $n$-gram model. They did not work on sequence recognition as they use the model for choreography synthesis. 
In general, activity recognition, in terms of sequence of atomic actions, has been explored widely. HMMs and {\em Dynamic Bayesian Networks} (DBNs) have been widely used for state model-based approaches (~\cite{park2004hierarchical, wu2016deep, natarajan2007coupled}), that represent a human activity as a model composed of a set of states. But, in the context of dance, multimodal analysis and recognition of dance sequence in terms of representative action elements are still open problems.

Given the challenges and the state-of-the-art solutions, we first extract the {\em Key Posture}s from a video using audio beats and {\em no-motion} information of the video. Next, we extract features from skeleton as well as RGB images of Kinect and use these features in three different classifiers for posture recognition. Finally, we design a recognizer for {\em Adavu}s using HMM. We also capture a rich data set for training and test, and annotate these with the help of {\em Bharatanatyam} experts. 
In this paper we address the basic three challenges of automatic analysis of dance using a multimodal framework. In this respect, our primary contributions are a multimodal framework for automatic extraction of key posture from a collection of dance sequences, recognizers of the key postures, and the capture of the formation rules of a {\em Bharatanayam Adavu} comprising of sequence of key postures.


The paper is organized as follows. We model the events for {\em Adavu}s and {\em Sollukattu}s in Section~\ref{sec:characterize}. Section~\ref{sec:data_set_annotation} presents the data set and annotation. Extraction of {\em Key Frames} is discussed in Section~\ref{sec:segmentation}. Section~\ref{sec:posture_recognizer} presents three recognizers for {\em Key Posture}s. {\em Adavu}s are recognized in Section~\ref{sec:adavu_recognizer}. Finally, we conclude in Section~\ref{sec:conclusion}.

\section{Event Models of {\em Bharatanatyam}s\label{sec:characterize}}
\begin{figure*}[!ht]
\centering
\includegraphics[width=14cm]{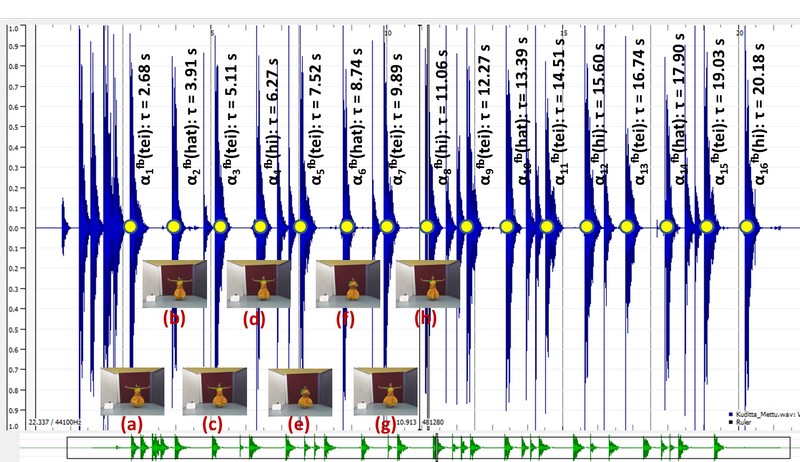}
\caption{Occurrence of {\em Key Posture}s in sync with the beat positions in {\em Kuditta Mettu Adavu}}
\label{fig:beat_kp}
\end{figure*}

Like in most dance forms, a {\em Bharatanatyam} dancer performs in sync with a specific form of  structured rhythmic music, called {\em Sollukattu}. 
Therefore, a {\em Bharatanatyam Adavu} consists of:

\begin{enumerate}
\item {\bf Audio Stream} or {\em Sollakattu} is generated by instrumental strikes along with vocal utterance. It is characterized by:
\begin{enumerate}
\item {\bf Beats}: A {\em Full Beat} or $1$-beat defines the basic unit of time –- an instance on the timescale generated by instrumental strikes. Elapsed time between two $1$-beats is called the {\em tempo period}. At times, in a {\em Sollukattu}, when the strike occurs at half the time or at one fourth of the time of tempo period, we call it the $\frac{1}{2}$-beat or $\frac{1}{4}$-beat respectively. 
\item {\bf Bols}: A {\em Bol} or vocal utterance {\em may} accompany a $1$-beat, $\frac{1}{2}$-beat or $\frac{1}{4}$-beat. 
\end{enumerate}

\item {\bf Video Stream} comprises frames of one of the following types:
\begin{enumerate}
\item {\bf {\em K-frame}s or Key Frames}: These frames contain {\em Key Posture}s where the dancer {\em holds} (remains momentarily stationary) the Posture. 

\item {\bf {\em T-frame}s of Transition Frame}: These are transition frames between two {\em K-frames} while the dancer is rapidly changing posture to assume the next {\em Key Posture} from the previous one. 
\end{enumerate} 

\item {\bf Synchronization}: Postures of an {\em Adavu} are performed in sync among themselves and in sync with the rhythm of the music. A {\em Bharatanatyam} dancer typically assumes a (momentarily stationary) {\em Key Posture} at a beat and makes a transition of posture between {\em Key Posture}s of two consecutive beats. In Figure~\ref{fig:beat_kp}, we show the {\em Key Posture}s for {\em Kuditta Mettu Adavu} using {\em Kuditta Mettu Sollukattu}.
\end{enumerate}

To formally characterize a {\em Sollukattu} and an {\em Adavu}, we define a set of audio, video and sync events for the above streams of information. The events play a critical role in maintaining the temporal consistency of dance. 

\subsection{Events of {\em Adavu}s}
An {\em Event} denotes the occurrence of a {\em Causal Activity} in the audio or the video stream of an {\em Adavu}. Further, sync events are defined between multiple events based on temporal constraints. 
An event is described by:

\begin{enumerate}
\item {\em Category}: Source of event -- {\em Audio}, {\em Video} or {\em Sync}.

\item {\em Type}: Nature of the causal activity of an event.

\item {\em Time-stamp / range}: The time of occurrence of the causal activity of the event. This is the elapsed time from the beginning of the stream and is marked by a function $\tau(.)$. Often a causal activity may spread over an interval $[\tau_s, \tau_e]$ which will be associated with the event. For video events, we use range of video frame numbers $[\eta_s, \eta_e]$ as the temporal interval. With a fixed frame rate of 30 fps in the video, we interchangeably use $[\tau_s, \tau_e]$ or $[\eta_s, \eta_e]$ for an event.

\item {\em Label}: One or more optional labels may be attached to an event annotating details for the causal activity.
 
\item {\em ID}: Every instance of an event in a stream is distinguishable. These are sequentially numbered in the temporal order of their occurrence. 
\end{enumerate}

The events are list in Table~\ref{tbl:events} with their characteristics. 
We locate and extract these events from the Kinect video to identify and segment various auditory (beats and {\em bol}s) and visual (postures) items for analysis and to reconstruct sequence information for in-depth structural understanding.

\begin{table}[!ht]
\caption{List of Events of {\em Adavu}s\label{tbl:events}}
\centering
\begin{small}
\begin{tabular}{|l|l|p{3cm}|p{2cm}|} \hline
\multicolumn{1}{|c|}{\bf Event} & \multicolumn{1}{c|}{\bf Event} & \multicolumn{1}{c|}{\bf Event} & \multicolumn{1}{c|}{\bf Event} \\
\multicolumn{1}{|c|}{\bf Category} & \multicolumn{1}{c|}{\bf Type} & \multicolumn{1}{c|}{\bf Description} & \multicolumn{1}{c|}{\bf Label} \\ \hline \hline
Audio & $\alpha^{fb}$ & Full- / $1$-beat$^1$ with {\em bol} & {\em bol}$^2$\\
Audio & $\alpha^{hb}$ & Half- / $\frac{1}{2}$-beat$^3$ with {\em bol} & {\em bol} \\
Audio & $\alpha^{qb}$ & $\frac{1}{4}$-beat$^4$ with {\em bol} & {\em bol} \\
Audio & $\alpha^{fn}$ & $1$-beat having no {\em bol}&  stick-beat$^5$ ($\bot$)   \\
Audio & $\alpha^{hn}$ & $\frac{1}{2}$-beat having no {\em bol}&  stick-beat ($\bot$)   \\
Audio & $\alpha^{qn}$ & $\frac{1}{4}$-beat having no {\em bol}&  stick-beat ($\bot$)   \\
Audio & $\beta$ & {\em bol} is vocalized & {\em bol} \\ \hline \hline

Video & $\nu^{nm}$ & No-motion$^6$ & Frame-Range$^7$, {\em Key Posture}$^{8}$ \\ 
Video & $\nu^{tr}$ & Transition Motion$^{9}$ & Frame-Range  \\  \hline \hline

Sync & $\psi^{fb}$ & No-motion @ $1$-beat$^{10}$ & {\em Key Posture} \\ 
Sync & $\psi^{hb}$ & No-motion @ $\frac{1}{2}$-beat$^{11}$ & {\em Key Posture} \\ \hline
\multicolumn{4}{l}{} \\
\end{tabular}

\begin{tabular}{rp{7.9cm}}
1: & A beat, often referred to as full-beat or $1$-beat, is the basic unit of time -- an instance on the timescale \\
2: & {\em bol}s accompany some beats ($1$-, $\frac{1}{2}$- or $\frac{1}{4}$-)\\
3: & A $\frac{1}{2}$-beat is a soft strike at the middle of a $1$-beat to $1$-beat gap or tempo period \\
4: & A $\frac{1}{4}$-beat is a soft strike at the middle of a $1$-beat to $\frac{1}{2}$-beat or a $\frac{1}{2}$-beat to $1$-beat gap  \\
5: & A stick-beat ($\bot$) has only beating and no {\em bol} \\ 
6: & Frames over which the dancer does not move ({\em Key Posture}) \\
7: & Sequence of consecutive frames over which the events spreads \\ 
8: & A {\em Key Posture} is a well-defined and stationery posture \\
9: & Transition motion to change from one {\em Key Posture} to the next \\
10: & $\alpha^{fb}$ and $\nu^{nm}$ in sync. That is, $\tau(\alpha^{fb}) \cap \tau(\nu^{nm}) \neq \phi$ \\ 
11: & $\alpha^{hb}$ and $\nu^{nm}$ in sync. That is, $\tau(\alpha^{hb}) \cap \tau(\nu^{nm}) \neq \phi$ \\ \hline
\end{tabular}
\end{small}
\end{table}

\section{Data Set and Annotation\label{sec:data_set_annotation}}
No data set for {\em Bharatanatyam Adavu}s is available for research. Hence, we start by recording a data set and then annotate them with the help of {\em Bharatanatyam} experts. A part of the data set has been published as \href{http://hci.cse.iitkgp.ac.in/}{\em Annotated Bharatanatyam Data Set}~\cite{mallick2017annotatedData} for reference and use by researchers.
\subsection{Data Set}
In an acoustic studio, we first record 6 sets (marked as SR1 through SR6) of 23 {\em Sollukattu}s using a {\em Zoom H2N Portable Handy Recorder}. At the time of recording the {\em Adavu}s, SR1 set is played back to the dancer. 
Next we record 10 sets (marked as AR1 through AR10) of 58 {\em Adavu}s with {\em Kinect XBox 360 (Kinect 1.0)} using {\em nuiCapture} software. These have been performed by 9 qualified dancers for one or more cycles and for one or more times. 

In this paper, we use on 8 variants of {\em Natta Adavu}s (named {\em Natta 1} through {\em Natta 8}) for posture and {\em Adavu} recognition. For this 
we choose AR1--AR3 and AR6--AR7 data sets (Table~\ref{tbl:Natta_Adavu_Data_Set}) based on the availability of completed annotation and mix of the female (AR1--AR3) and male (AR6--AR7) dancers. 
 
\begin{table}[!ht]
\caption{{\em Natta Adavu} Data Set\label{tbl:Natta_Adavu_Data_Set}}
\centering
\begin{tabular}{|l|l|r|r|r|} \hline 
\multicolumn{1}{|c}{\bf Recording} & 
\multicolumn{1}{|c}{\bf Dancer}  & 
\multicolumn{1}{|c}{\bf \# of} &
\multicolumn{1}{|c}{\bf Performances} &
\multicolumn{1}{|c|}{\bf \# of} \\ 

\multicolumn{1}{|c}{\bf Set \#} & 
\multicolumn{1}{|c}{\bf  ID}  & 
\multicolumn{1}{|c}{\bf Cycles}	&
\multicolumn{1}{|c}{\bf per {\em Adavu}}	&
\multicolumn{1}{|c|}{\bf Recordings} \\ \hline \hline
AR1	&	D1	&	1	&	1	&	8 * 1 = 8  \\ \hline
AR2	&	D2	&	1	&	1	&	8 * 1 = 8  \\ \hline
AR3	&	D3	&	1	&	1	&	8 * 1 = 8  \\ \hline
AR6	&	D6	&	4	&	3	&	8 * 3 = 24 \\ \hline
AR7	&	D7	&	4	&	3	&	8 * 3 = 24 \\ \hline
\multicolumn{4}{l}{ } \\
\multicolumn{5}{p{8cm}}{\footnotesize Each set has performances for 8 variants of {\em Natta Adavu}s} \\
\multicolumn{4}{l}{ } \\
\multicolumn{4}{l}{\bf Total Recordings} & \multicolumn{1}{r}{\bf 72} \\
\end{tabular}
\end{table}

\subsection{Annotation}
\begin{figure}[!ht]
\centering
	\subfigure[Natta1P1]
		{\includegraphics[width=3cm]{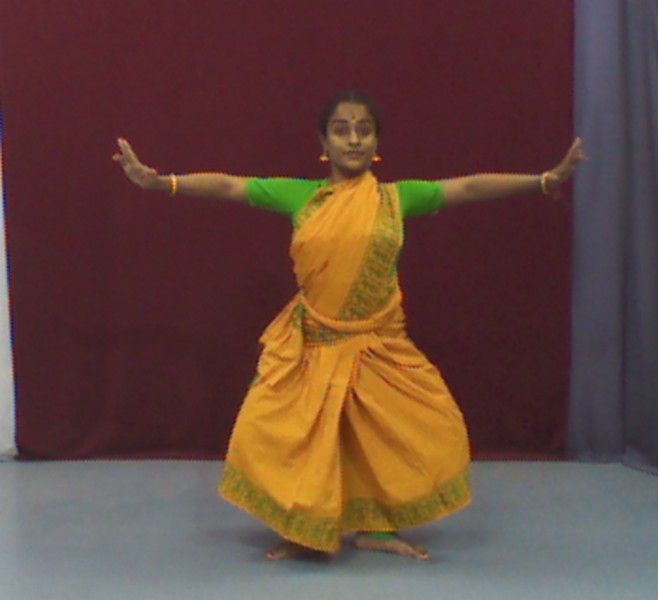}}
	\hspace{0.05 cm}
	\subfigure[Natta1P2]
		{\includegraphics[width=3cm]{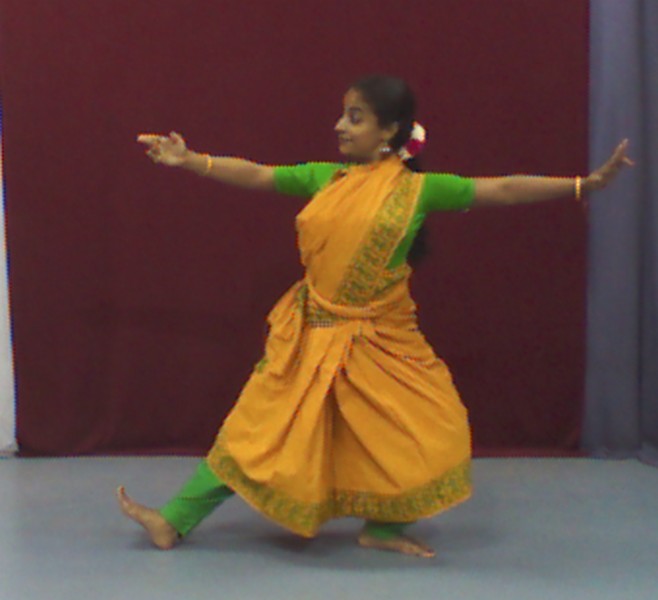}}
	\hspace{0.05 cm}
	\subfigure[Natta1P1]
		{\includegraphics[width=3cm]{Images/Natta1P1}}
	\hspace{0.05 cm}
	\subfigure[Natta1P3]
		{\includegraphics[width=3cm]{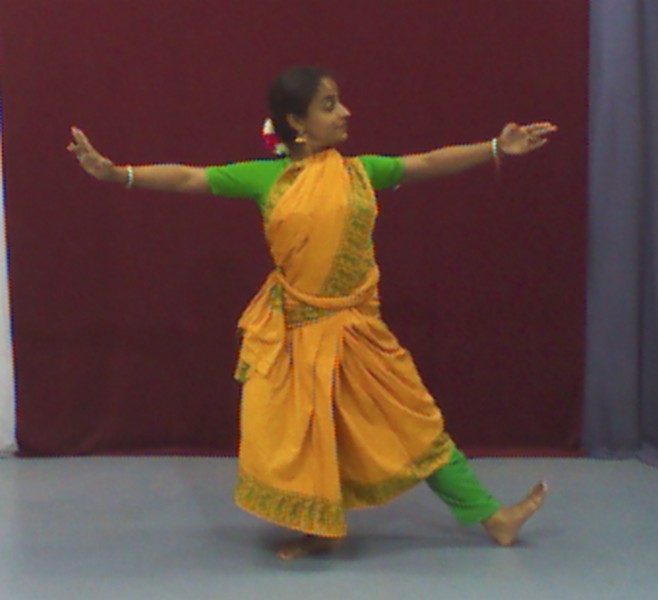}}
\caption{Cyclic occurrences of {\em Key Posture}s in {\em Natta 1 Adavu}}
\label{fig:cycle_natta}
\end{figure}

With the help of {\em Bharatanatyam} experts, we have annotated the data set of {\em Natta Adavu}s (Table~\ref{tbl:Natta_Adavu_Data_Set}). The steps of annotation are:
\begin{enumerate}
\item Segment the video in alternating sequences of {\em K-frames} and {\em T-frames}. In a sequence of {\em K-frames} the dancer is almost stationery. Intervening frames with motion form sequence of {\em T-frames}. For every sequence the range of RGB frame numbers is noted. The clues from the audio (beats and {\em bol}s) are also used in the segmentation process because every {\em K-frame} must have been triggered by a beat having an optional {\em bol}.


\item Select one representative {\em K-frame} each from every sequence of {\em K-frames}. 
Every {\em K-frame} has a {\em Key Posture}.


\item Annotate the {\em Key Posture} in every {\em K-frame}.

\item Record the annotations in a file.
\end{enumerate}
For example, in {\em Natta 1 Adavu}, the dancer cycles through 4 {\em Key Posture}s (Figure~\ref{fig:cycle_natta}) using {\em Natta Sollukattu}. This is annotated in Table~\ref{tbl:annotated_data}. The range of {\em K-frame}s for every {\em Key Posture} is noted with the beats and {\em bol}s while {\em T-frame}s occur in-between {\em K-frame}s.

\begin{table}[!ht]
\centering
\caption{Annotations of a video of {\em Natta 1 Adavu}\label{tbl:annotated_data}}
\begin{small}
\begin{tabular}{|l|r|r|r|l|} \hline
\multicolumn{1}{|c}{\bf Posture}	&	\multicolumn{1}{|c}{\bf Start}	&	\multicolumn{1}{|c}{\bf End}	&	\multicolumn{1}{|c}{\bf Beat}	&	\multicolumn{1}{|c|}{\bf Bols}	\\ 
\multicolumn{1}{|c}{\bf Name}	&	\multicolumn{1}{|c}{\bf Frame}	&	\multicolumn{1}{|c}{\bf Frame}	&	\multicolumn{1}{|c}{\bf Number}	&	\multicolumn{1}{|c|}{\bf }	\\ 
\multicolumn{1}{|c}{\bf (a)}	&	\multicolumn{1}{|c}{\bf (b)}	&	\multicolumn{1}{|c}{\bf (c)}	&	\multicolumn{1}{|c}{\bf (d)}	&	\multicolumn{1}{|c|}{\bf (e)}	\\ \hline \hline
Natta1P1 ({\bf C01})	&	70	&	89	&	0	&	No Bol	\\ \hline
Natta1P2 ({\bf C02})	&	101	&	134	&	1	&	{\em tei yum}	\\ \hline
Natta1P1 ({\bf C01})	&	144	&	174	&	2	&	{\em tat tat}	\\ \hline
Natta1P3 ({\bf C03})	&	189	&	218	&	3	&	{\em tei yum}	\\ \hline
Natta1P1 ({\bf C01})	&	231	&	261	&	4	&	{\em ta}	\\ \hline
\multicolumn{5}{l}{} \\
\multicolumn{5}{l}{\footnotesize Class names / Posture IDs are also marked from Figure~\ref{fig:natta_adavus_postures}}
\end{tabular}
\end{small}
\end{table}

\begin{figure}[!ht]
\centering
\begin{scriptsize}
\begin{tabular}{cccccc}
\includegraphics[width=2.5cm]{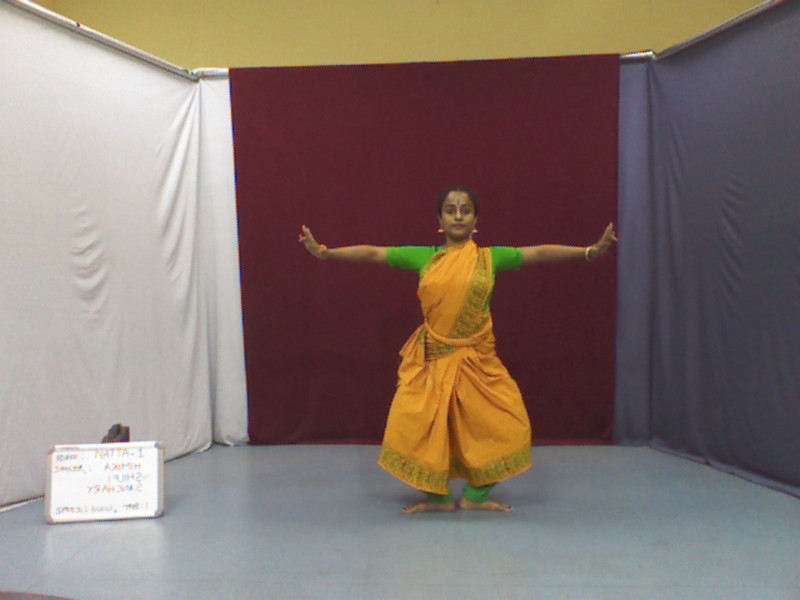} & \includegraphics[width=2.5cm]{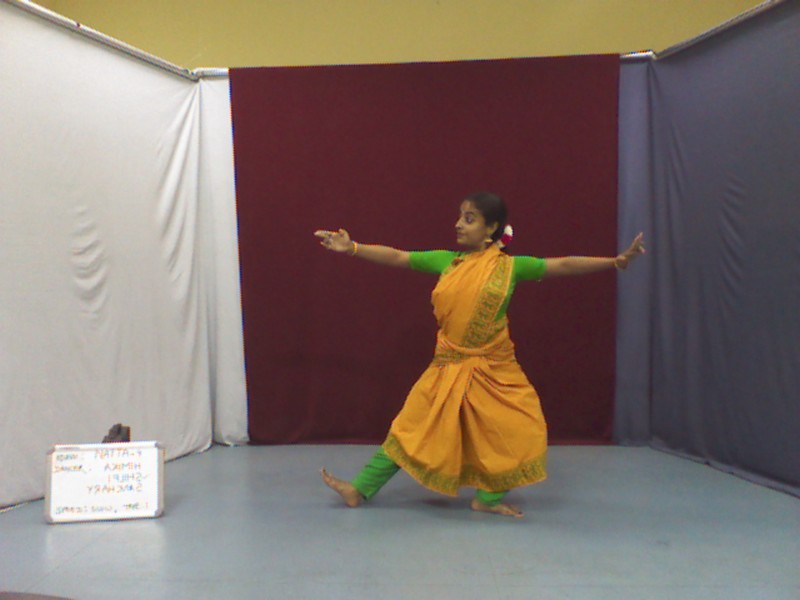} & \includegraphics[width=2.5cm]{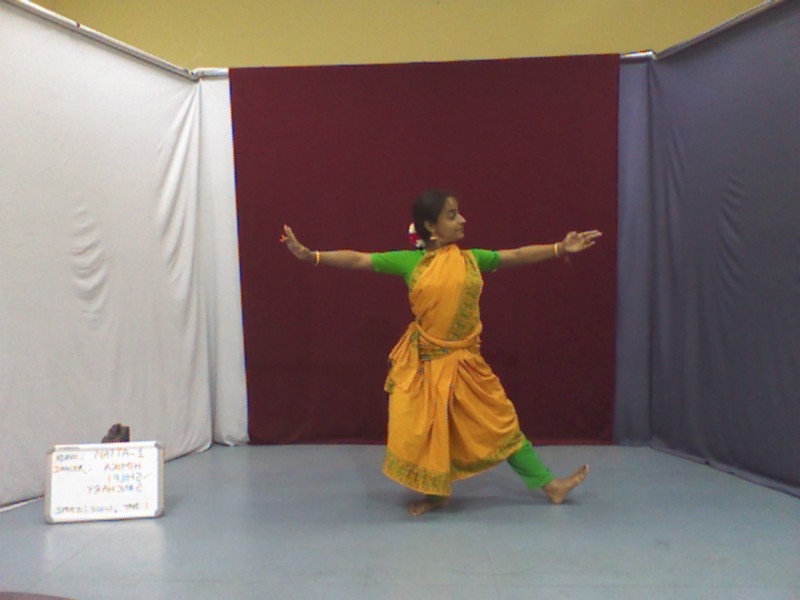} &
\includegraphics[width=2.5cm]{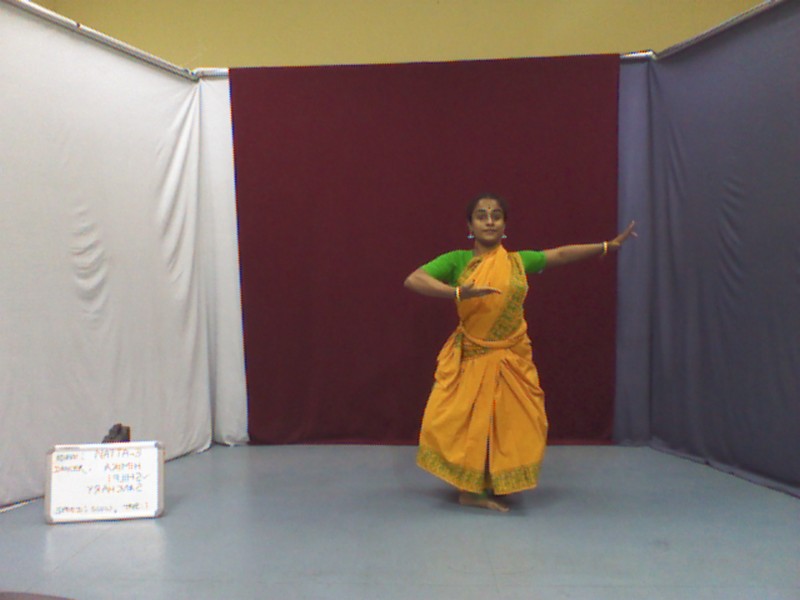} & \includegraphics[width=2.5cm]{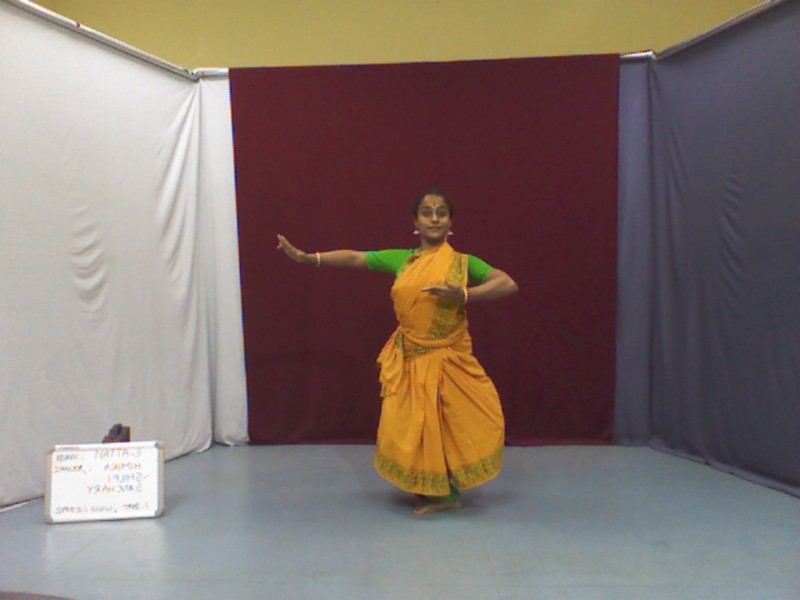} & \includegraphics[width=2.5cm]{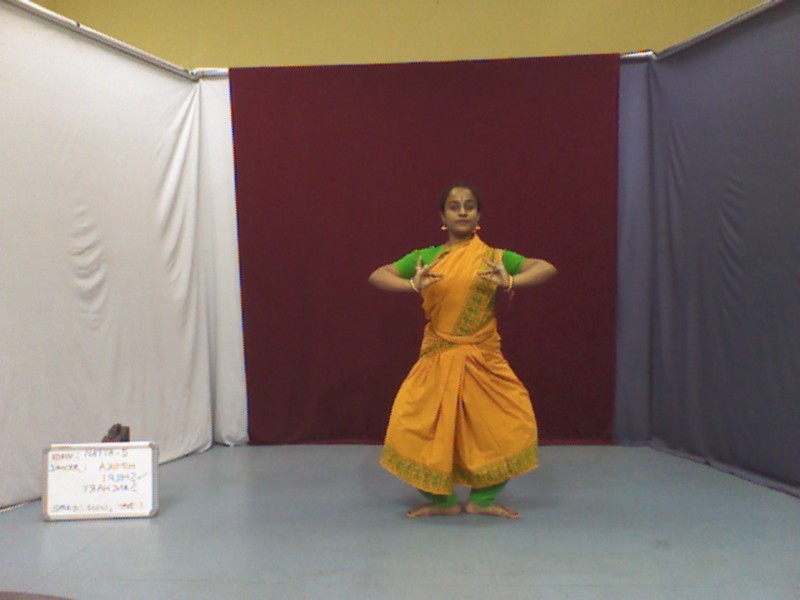} \\
{\bf C01} & {\bf C02} & {\bf C03} & {\bf C04} & {\bf C05} & {\bf C06}\\

\includegraphics[width=2.5cm]{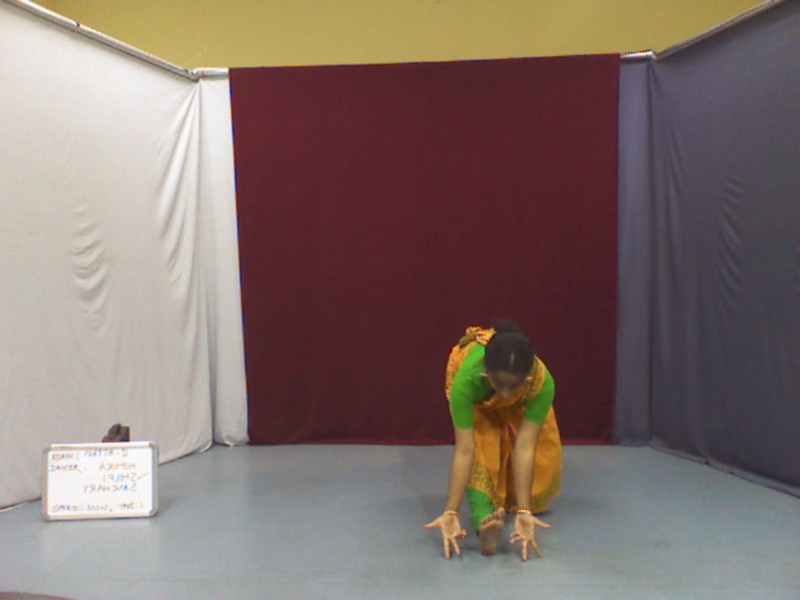} & \includegraphics[width=2.5cm]{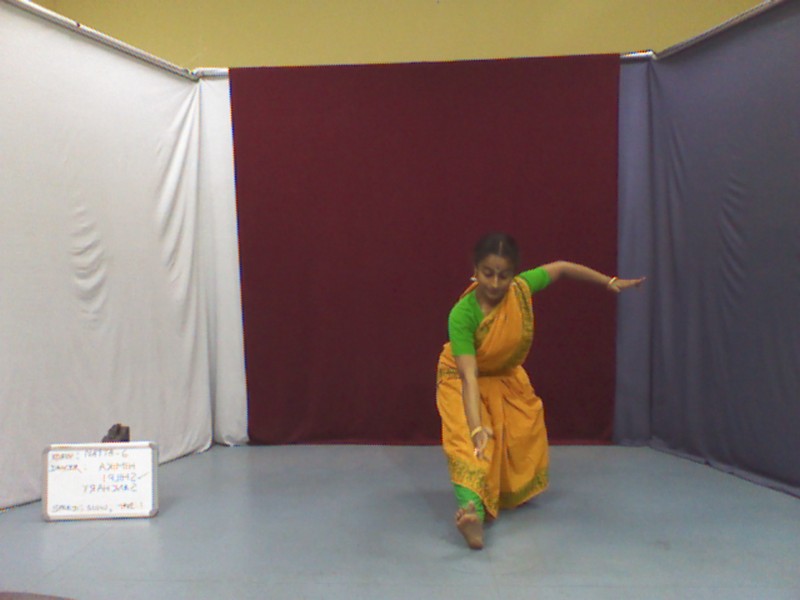} & \includegraphics[width=2.5cm]{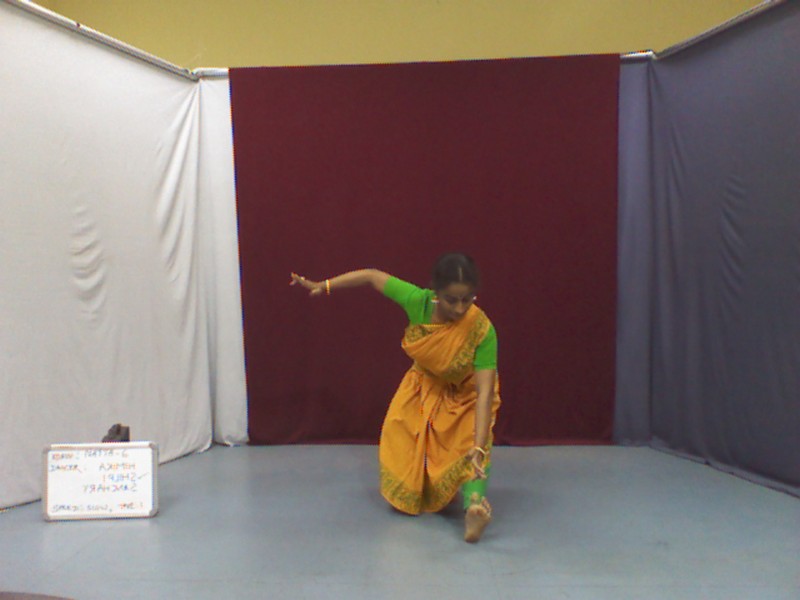} &
\includegraphics[width=2.5cm]{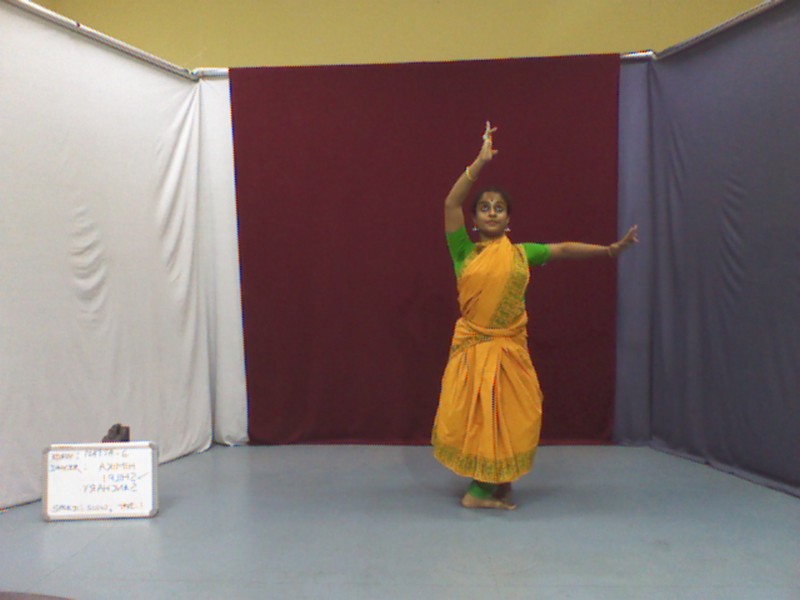} & \includegraphics[width=2.5cm]{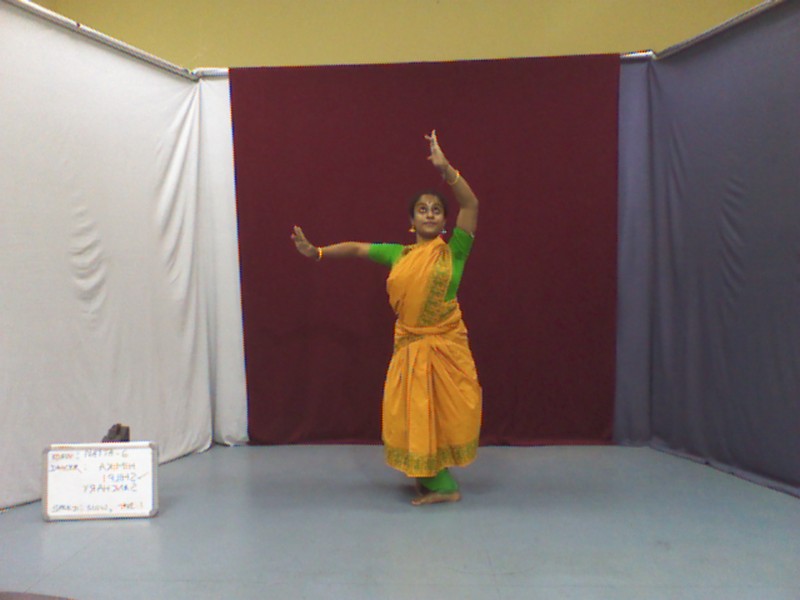} & \includegraphics[width=2.5cm]{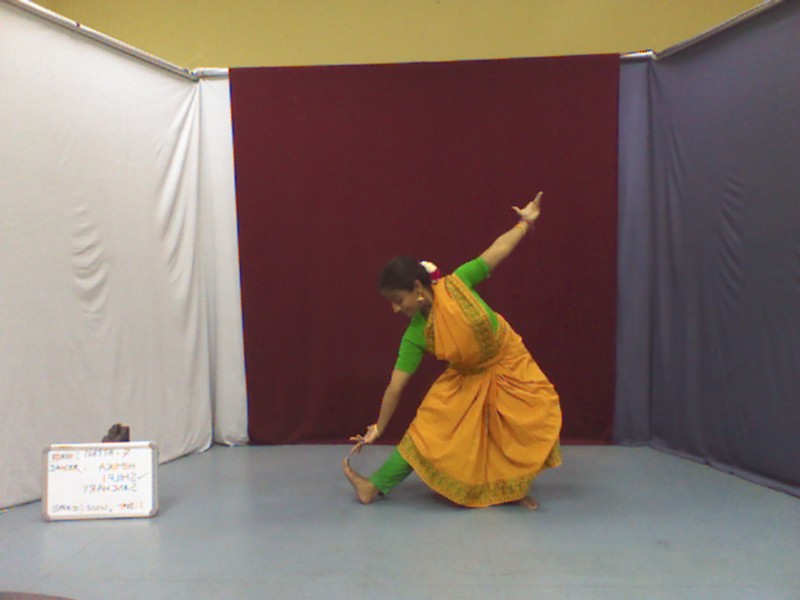} \\
{\bf C07} & {\bf C08} & {\bf C09} & {\bf C10} & {\bf C11} & {\bf C12} \\

\includegraphics[width=2.5cm]{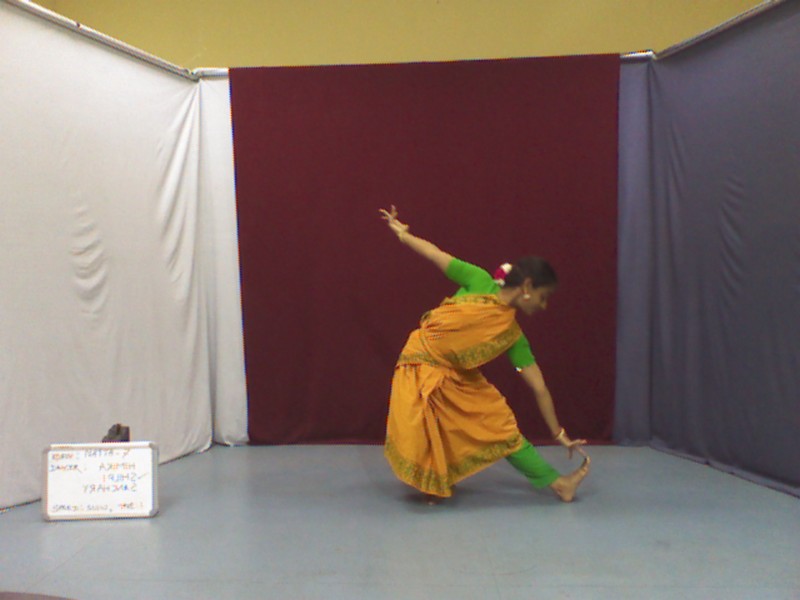} & \includegraphics[width=2.5cm]{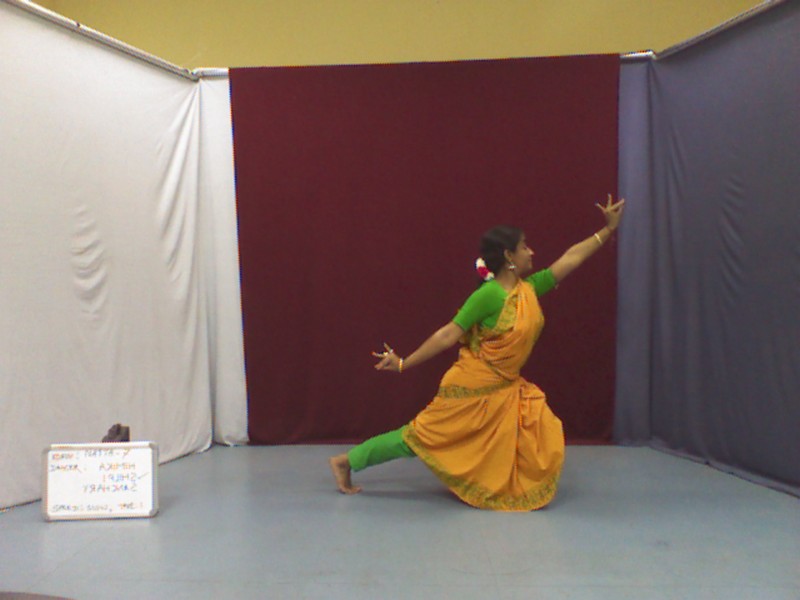} & \includegraphics[width=2.5cm]{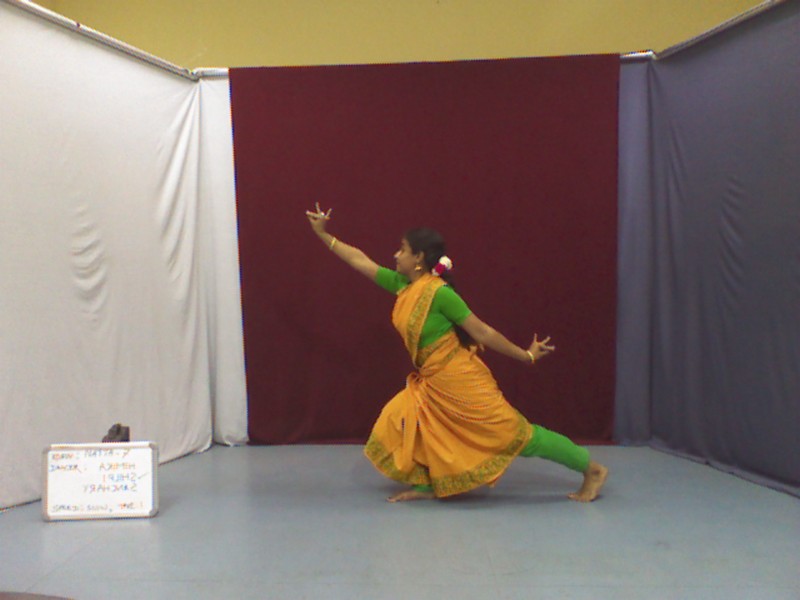} & 
\includegraphics[width=2.5cm]{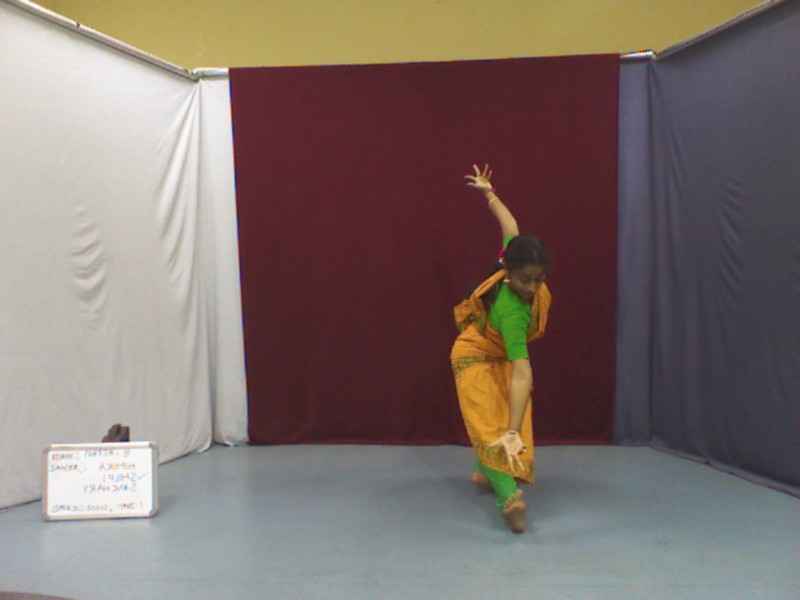} & \includegraphics[width=2.5cm]{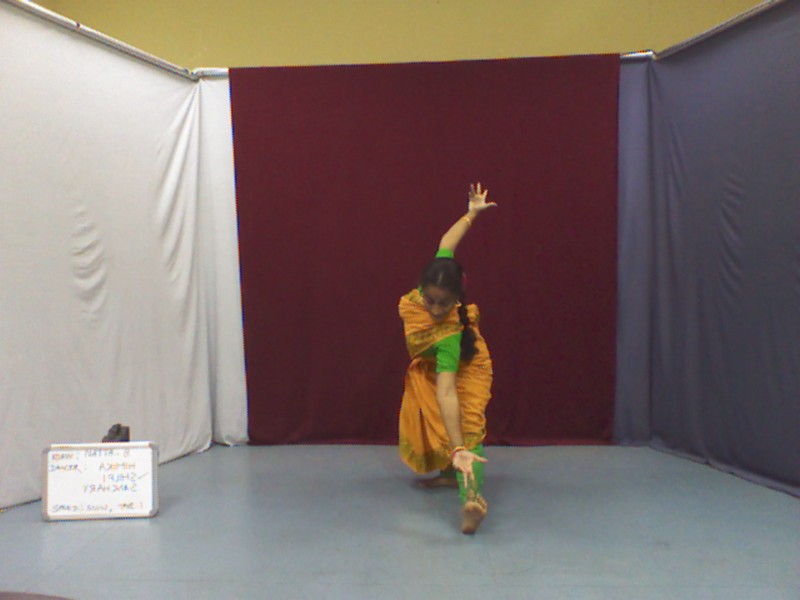} & \includegraphics[width=2.5cm]{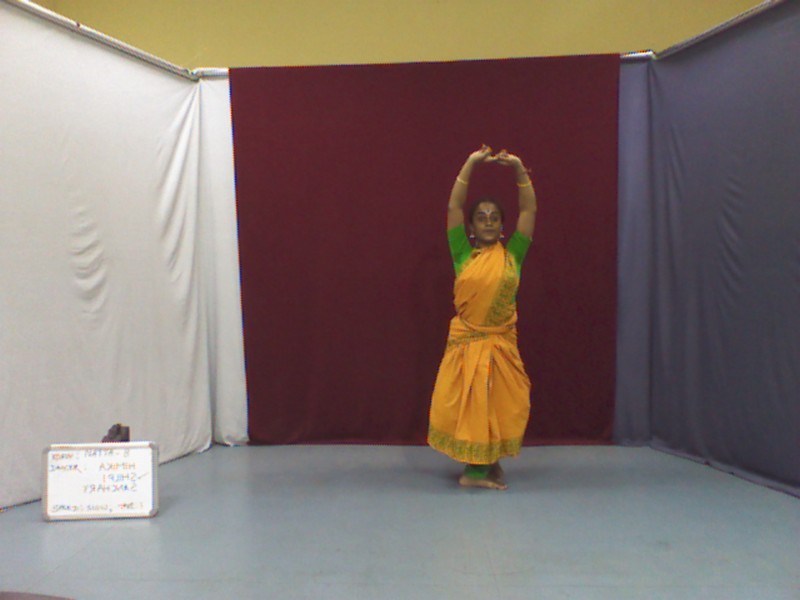} \\
{\bf C13} & {\bf C14} & {\bf C15} & {\bf C16} & {\bf C17} & {\bf C18} \\

\includegraphics[width=2.5cm]{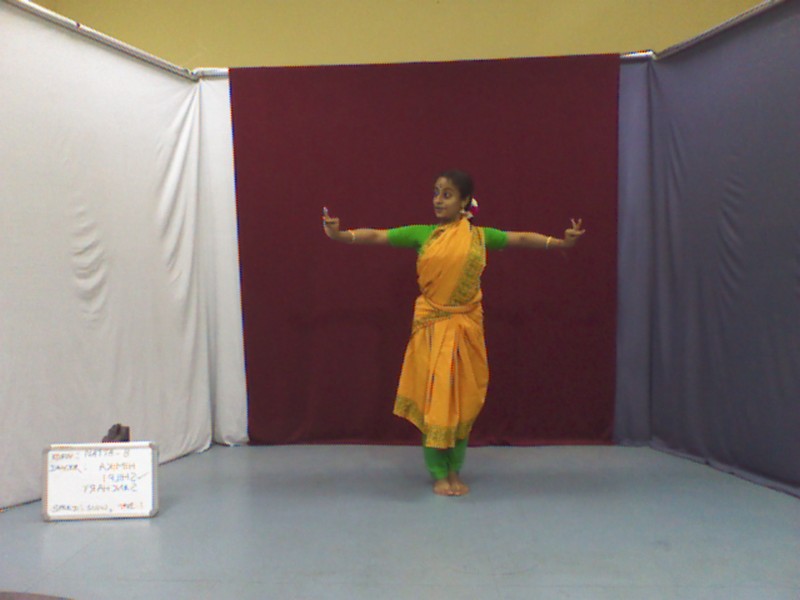} & \includegraphics[width=2.5cm]{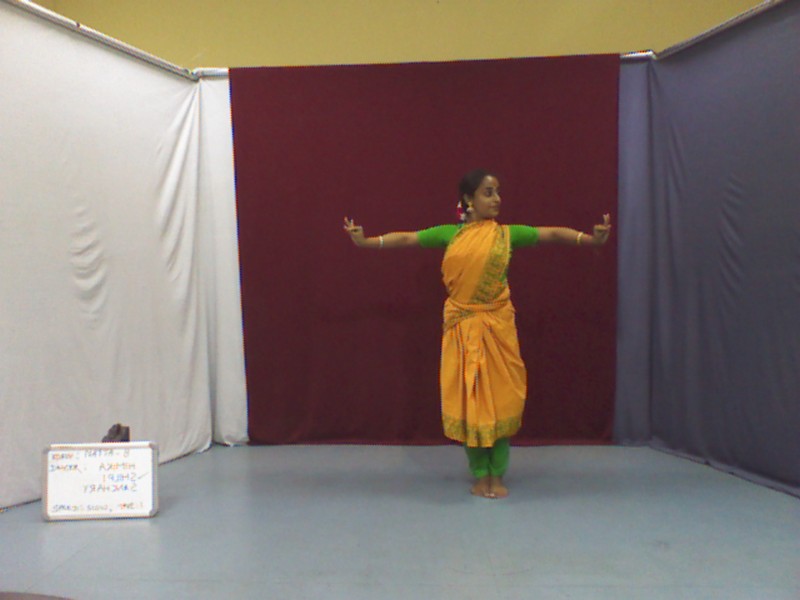} & \includegraphics[width=2.5cm]{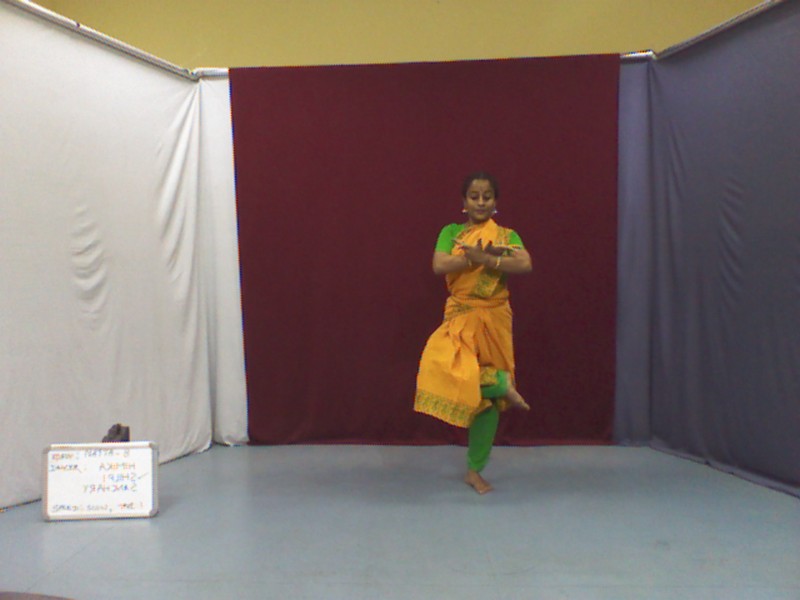} & 
\includegraphics[width=2.5cm]{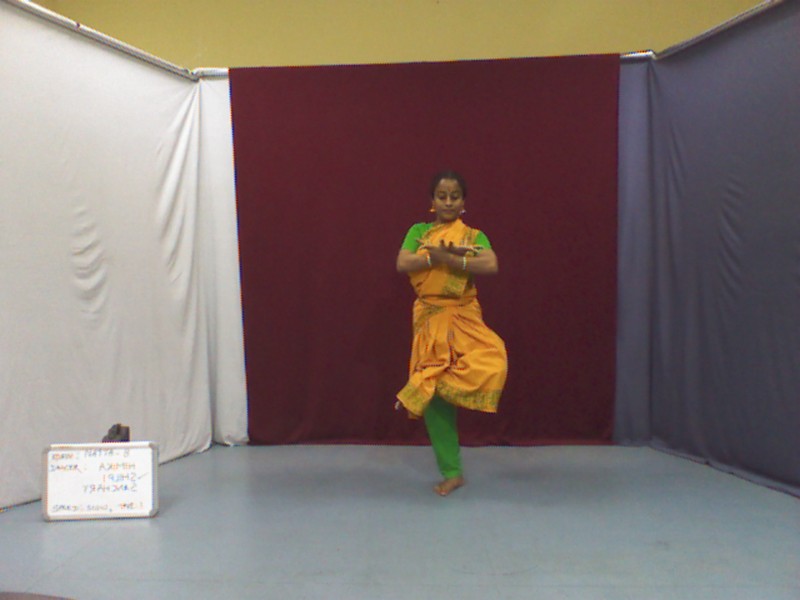} & \includegraphics[width=2.5cm]{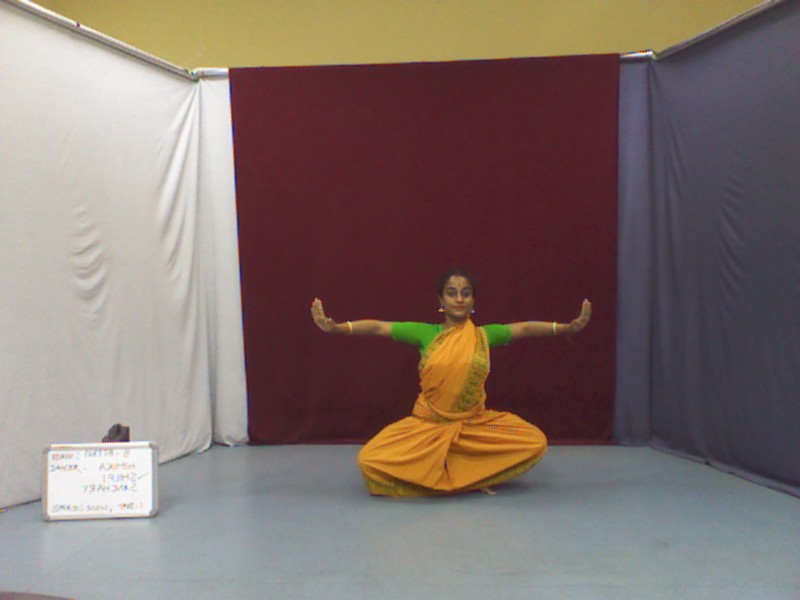} &\\
{\bf C19} & {\bf C20} & {\bf C21} & {\bf C22} & {\bf C23} & \\

\end{tabular}
\end{scriptsize}
\caption{23 {\em Key Posture}s of {\em Natta Adavu}s with Class / Posture IDs\label{fig:natta_adavus_postures}}
\end{figure}

Using the annotation we collect all {\em K-frame}s from the 72 videos (Table~\ref{tbl:Natta_Adavu_Data_Set}) to form our posture data set. We identify 23 unique {\em Key Posture}s (Figure~\ref{fig:natta_adavus_postures}) that occur in 8 {\em Natta Adavu}s. Complete annotation helps us to link every {\em K-frame} to one of the 23 classes. The number of samples for every class is shown in Table~\ref{tab:posture_data}. The data is skewed across classes due to the skew in the occurrence of the postures in the {\em Adavu}s.

\begin{table}[!ht]
\centering 
\caption{Data Set for Posture Recognition\label{tab:posture_data}}
\begin{tabular}{|l|r|r||l|r|r|} \hline
\multicolumn{1}{|c}{\bf Posture} & \multicolumn{1}{|c}{\bf Training} & \multicolumn{1}{|c}{\bf Test} & \multicolumn{1}{||c}{\bf Posture} & \multicolumn{1}{|c}{\bf Training} & \multicolumn{1}{|c|}{\bf Test} \\ 
\multicolumn{1}{|c}{\bf ID} & \multicolumn{1}{|c}{\bf data} & \multicolumn{1}{|c}{\bf data} & \multicolumn{1}{||c}{\bf ID} & \multicolumn{1}{|c}{\bf data} & \multicolumn{1}{|c|}{\bf data} \\ \hline \hline
{\bf C01} & 6154 & 1457 & {\bf C13} & 235 & 80\\
{\bf C02} & 3337 & 873	& {\bf C14} & 393 & 117\\
{\bf C03} & 3279 & 561 & {\bf C15} & 404 & 121\\
{\bf C04} & 1214 & 219 & {\bf C16} & 150 & 48\\
{\bf C05} & 1192 & 268 & {\bf C17} & 161 & 51\\
{\bf C06} & 1419 & 541 & {\bf C18} & 323 & 81\\
{\bf C07} & 1250 & 475 & {\bf C19} & 175 & 46\\
{\bf C08} & 284 & 112 & {\bf C20} & 168 & 43\\
{\bf C09} & 306 & 133 & {\bf C21} & 19 & 6\\
{\bf C10} & 397 & 162 & {\bf C22} & 21 & 6\\
{\bf C11} & 408 & 117 & {\bf C23} & 118 & 61\\
{\bf C12} & 229 & 84 & & & \\ \hline
 \multicolumn{6}{l}{} \\
 \multicolumn{6}{p{7.5cm}}{\footnotesize Numbers indicate the number of {\em K-frame}s in each class. Each {\em K-frame} is denoted by the frame number of the RGB frame in the video. The associated depth and skeleton frames are used as needed. Various position and formation information on body parts are available for every {\em K-frame} from annotation} \\
\end{tabular}
\end{table}

\section{Extraction of Key Frames\label{sec:segmentation}} 
The {\em K-frame}s are synchronized with music beats. Hence, we first work on beat marking and automatic annotation of 
{\em Sollukattu}~\cite{mallick2017Sollukattu}. We get the time-stamps of full- and half-beats ($\alpha^{fb}$ and $\alpha^{hb}$ events) with the corresponding {\em bol}s. Here, we intend to segment the video into {\em K-frame}s and {\em T-frame}s in an audio-guided segmentation using these events. 
However, while synchronizing audio and video events it should be noted that a {\em Key Posture}  occurs only at $\alpha^{fb}$ or $\alpha^{hb}$ positions, but every $\alpha^{fb}$ or $\alpha^{hb}$ does not necessarily associate with a key posture. For example, Figure~\ref{fig:beat_kp}  shows that there is no {\em Key Posture} at the $\alpha^{hb}$ position for this particular {\em Adavu}. Hence, after the audio-guided segmentation we get a set of {\em candidate} {\em K-frame}s only.

\begin{table*}
\centering
\caption{Predicates over events\label{tbl_predicates}}
\begin{tabular}{|llll|} \hline
Seq. of Frames from Audio Events & : &  \textcolor{blue}{$[\eta_s(\alpha_1),\eta_e(\alpha_1)], [\eta_s(\alpha_2),\eta_e(\alpha_2)], \cdots,  [\eta_s(\alpha_i),\eta_e(\alpha_i)]$} & (\myEqnCnt{eqn:audio_event})) \newline \\

& : & $\quad$ where \textcolor{blue}{$\forall i, i \geq 1$: $\alpha_i$ $\in$ \{$\alpha^{fb}$, $\alpha^{hb}$\}, $\eta_s(\alpha_i) < \eta_e(\alpha_i)$, } & \\

& : & $\quad\quad\quad$ \textcolor{blue}{ and $\eta_s(\alpha_{i+1}) > \eta_e(\alpha_i)$} & \\

Seq.\ of\ Frames\ from\ Video\ Events & : & \textcolor{magenta}{$[\eta_s(\nu^{nm}_1),\eta_e(\nu^{nm}_1)], [\eta_s(\nu^{tr}_2),\eta_e(\nu^{tr}_2)], [\eta_s(\nu^{nm}_3),\eta_e(\nu^{nm}_3)], $} & (\myEqnCnt{eqn:video_event})) \newline \\

 & : & $\quad$ \textcolor{magenta}{$ [\eta_s(\nu^{tr}_4),\eta_e(\nu^{tr}_4)], \cdots,$} &  \newline \\

& : & $\quad$ where \textcolor{magenta}{$\forall i, i \geq 1$: $\eta_s(\nu_{i+1}) = \eta_e(\nu_i) + 1$} & \\

Seq. of K-Frames & : & \textcolor{blue}{$[\eta_s(\psi_1),\eta_e(\psi_1)], [\eta_s(\psi_2),\eta_e(\psi_2)], \cdots, [\eta_s(\psi_i),\eta_e(\psi_i)], \forall i, i \geq 1$} & (\myEqnCnt{eqn:k_frames})) \newline \\

$\quad$No-Motion Predicate & : & \textcolor{blue}{$\quad$ $\exists j, j \geq i: [\eta_s(\psi_i),\eta_e(\psi_i)] = [\eta_s(\nu^{nm}_j),\eta_e(\nu^{nm}_j)]$} & (\myEqnCnt{eqn:no_motion})) \newline \\

$\quad$Overlap Predicate & : & \textcolor{blue}{$\quad$ $\exists k: [\eta_s(\alpha_k),\eta_e(\alpha_k)] \cap [\eta_s(\nu^{nm}_j),\eta_e(\nu^{nm}_j)] \neq \phi$} & (\myEqnCnt{eqn:overlap})) \newline \\

K-Frames Annotation & : & \textcolor{magenta}{$[\eta_s(a_1),\eta_e(a_1)], [\eta_s(a_2),\eta_e(a_2)], \cdots, [\eta_s(a_i),\eta_e(a_i)], \forall i, i \geq 1$} & (\myEqnCnt{eqn:annotation})) \newline \\

Validation Predicate & : & \textcolor{blue}{$\exists j: [\eta_s(\psi_i),\eta_e(\psi_i)] \cap [\eta_s(a_j),\eta_e(a_j)] \neq \phi$} & (\myEqnCnt{eqn:validate})) \\ \hline

\end{tabular}
\end{table*}

In Table~\ref{tbl_predicates}, we summarize the set of predicates that hold over various events and explain these here. First, using the audio events, the video is represented by predicates~(\ref{eqn:audio_event}).
We next perform a motion-guided segmentation, to classify every frame as no-motion or motion, to determine $\nu^{nm}$ and  $\nu^{tr}$ events  that  alternate as in  predicates~(\ref{eqn:video_event}). 
Finally, we combine the audio-guided and motion-guided segmentation results to determine the final set of {\em K-frame}s. This is represented as in predicates~(\ref{eqn:k_frames}), where $\psi_i$ is an $\psi^f$ or $\psi^h$ event and satisfy the predicates as in~(\ref{eqn:no_motion}) and~(\ref{eqn:overlap}) respectively. So we compute the set of ranges of {\em K-frame}s (predicates~(\ref{eqn:k_frames})) from the set of ranges of frames having audio events (as in predicates~(\ref{eqn:audio_event})) and the set of range of frames having video events (predicates~(\ref{eqn:video_event})). The range of a selected {\em K-frame} equals a range of frames having {\em no motion} (predicate in~(\ref{eqn:no_motion})) and has  overlap (full or partial) with the range of frames for an audio event (predicate in~(\ref{eqn:overlap})). Finally, the data is annotated for {\em K-frame}s as in predicates~(\ref{eqn:annotation}), where $a_i$ denotes $i^{th}$ annotation. One can now validate a frame as {\em K-frame} if it satisfies the validation predicate as in~(\ref{eqn:validate}). Otherwise, the validation fails.

\subsection{Motion-Guided Extraction of Key Frames}\label{sec:no_motion_detection}
We try to detect {\em no-motion}\footnote{Actually, slow or low motion in the video as cutoff by a threshold} ($\nu^{nm}$ events) in the video. Given that $\nu^{nm}$ and $\nu^{tr}$ must alternate in the video, we then deduce the $\nu^{tr}$ events. 
A {\em no-motion} event can be detected from Frame Differences in the RGB stream. This is achieved by converting RGB frame data to gray-scale and then computing the frame difference (pixel-wise absolute difference of intensity values) between consecutive frames. Every pixel of the difference frames is binarized into {\em motion} (marked as 1) and {\em no-motion} (marked as 0) based on a threshold $Th_{pix}$ (higher the value, more is the likelihood of motion). We sum the motion values (number of 1s in a frame) in every binarized frame and use another threshold $Th_{frm}$ to mark the frame as a {\em motion} frame or a {\em no-motion} frame and generate a sequence as in predicates~(\ref{eqn:video_event}). 

\begin{algorithm}[!ht]
  \caption{: No-motion detection}\label{algo_elmd}
  \begin{algorithmic}[1]
  \State {\bf Inputs:} $frame_i$ = Grayscale Image of $i^{th}$ frame of the video (reference frame), $frame_{i+1}$ = Grayscale Image of $i^{th}$ frame of the video (target frame), $Th_{pix}$, $Th_{frm}$
  \State {\bf Output:}  $0$ for no-motion otherwise $1$ 
\For{ $r = 1:number\_of\_rows$}
	\For{ $c = 1:number\_of\_columns$}
		\If{$frame_{i+1}(r, c) - frame_i(r, c) >  Th_{pix} $}
			\State $frame_{i+1}(r, c) = 1 $;
        \Else
            \State $frame_{i+1}(r, c) = 0 $;
        \EndIf
    \EndFor
\EndFor
\State $ s = sum(frame_{i+1}(r, c)) $;
\If{$s >  Th_{frm}$}
	\State return 1;
\Else
    \State return 0;
\EndIf
\end{algorithmic}
\end{algorithm}



The threshold $Th_{pix}$ is set at slightly high value of 50 (that is, about 20\% of the maximum intensity 255) to avoid random variations of intensity values and noise being falsely detected as motion at a pixel. To decide on the threshold $Th_{frm}$, we note that motion in a dance frame can be caused only by the movement of the dancer (background and all other objects in the field of view are stationary). Given the setup for imaging, we estimate (over 1000 images) that the dancer typically occupies less than 10\% of the whole frame. Further, the dancer usually moves only the major body parts (legs and hands) which comprises less than 50\% of her body. 
To mandate a strict stationary position (as is required for a {\em Key Posture}), we limit spurious motion to about 1\% only. Hence for a whole frame, we expect that not more than 10\% * 50\% * 1\% = 0.05\% of the pixels to have motion in a {\em K-frame}. Given a frame size of 640 $\times$ 480 pixels, this translates to $\approx$ 150 pixels. After some experiments with the recorded data around this threshold, we finally set $Th_{frm} = 100$ pixels.

\begin{table*}[!ht]
\caption{Results of Extraction of Key Frames in {\em Natta 3 Adavu}\label{tab:kp_natta3}}
\centering
\begin{scriptsize}
\begin{tabular}{|r||r|r|l|l||r|r||r|r||r|r|} \hline
\multicolumn{1}{|c||}{\bf Line} & \multicolumn{4}{c||}{\bf Audio Beats Mapped to} & \multicolumn{2}{c||}{\bf {\em No Motion}} & \multicolumn{2}{c||}{\bf Overlap of Beat} & \multicolumn{2}{c|}{\bf Annotation of} \\ 
\multicolumn{1}{|c||}{\bf \#} & \multicolumn{4}{c||}{\bf Video Frames} & \multicolumn{2}{c||}{\bf from Video} & \multicolumn{2}{c||}{\bf \& {\em No Motion}} & \multicolumn{2}{c|}{\bf {\em Key Posture}s} \\ \cline{2-11} 
\multicolumn{1}{|c||}{} & \multicolumn{1}{c|}{\bf Start} & \multicolumn{1}{c|}{\bf End} & \multicolumn{1}{c|}{\bf {\em Bol}} & \multicolumn{1}{c||}{\bf Beat} & \multicolumn{1}{c|}{\bf Start} & \multicolumn{1}{c||}{\bf End} & \multicolumn{1}{c|}{\bf Start} & \multicolumn{1}{c||}{\bf End} & \multicolumn{1}{c|}{\bf Start} & \multicolumn{1}{c|}{\bf End} \\ 
\multicolumn{1}{|c||}{} & \multicolumn{1}{c|}{\bf Frame} & \multicolumn{1}{c|}{\bf Frame} & \multicolumn{1}{c|}{\bf } & \multicolumn{1}{c||}{\bf Info} & \multicolumn{1}{c|}{\bf Frame} & \multicolumn{1}{c||}{\bf Frame} & \multicolumn{1}{c|}{\bf Frame} & \multicolumn{1}{c||}{\bf Frame} & \multicolumn{1}{c|}{\bf Frame} & \multicolumn{1}{c|}{\bf Frame} \\ 
\multicolumn{1}{|c||}{} & \multicolumn{1}{c|}{\bf $\eta_s(\alpha^f)$} & \multicolumn{1}{c|}{\bf $\eta_e(\alpha^f)$} & \multicolumn{1}{c|}{\bf } & \multicolumn{1}{c||}{\bf } & \multicolumn{1}{c|}{\bf $\eta_s(\nu^{nm})$} & \multicolumn{1}{c||}{\bf $\eta_e(\nu^{nm})$} & \multicolumn{1}{c|}{\bf $\eta_s(\psi^f)$} & \multicolumn{1}{c||}{\bf $\eta_e(\psi^f)$} & \multicolumn{1}{c|}{\bf $\eta_s$} & \multicolumn{1}{c|}{\bf $\eta_e$} \\ \cline{2-11} 
\multicolumn{1}{|c||}{} & \multicolumn{1}{c|}{(a)} & \multicolumn{1}{c|}{(b)} & \multicolumn{1}{c|}{(c)} & \multicolumn{1}{c||}{(d)} & \multicolumn{1}{c|}{(e)} & \multicolumn{1}{c||}{(f)} & \multicolumn{1}{c|}{(g)} & \multicolumn{1}{c||}{(h)} & \multicolumn{1}{c|}{(i)} & \multicolumn{1}{c|}{(j)} \\ \hline\hline
1 & & & & & 3 & 94 & & & & \\ \hline
2 & 98 & 109 & {\em tei yum} & B-HB & 108 & 132 & 108 & 132 & 104 & 135 \\ \hline
3 & 143 & 173 & {\em tat tat} & B-HB & 142 & 161 & 142 & 161 & 147 & 173 \\ \hline
4 & & & & & 163 & 186 & 163 & 186 & & \\ \hline
5 & 186 & 220 & {\em tei yum} & B-HB & 188 & 217 & 188 & 217 & 190 & 216 \\ \hline
6 & 230 & 242 & {\em ta} & B & 223 & 256 & 223 & 256 & 232 & 259 \\ \hline
7 & 273 & 306 & {\em tei yum} & B-HB & 279 & 305 & 279 & 305 & 274 & 303 \\ \hline
8 & 314 & 344 & {\em tat tat} & B-HB & 322 & 344 & 322 & 344 & 318 & 341 \\ \hline
9 & 357 & 390 & {\em tei yum} & B-HB & 363 & 391 & 363 & 391 & 363 & 390 \\ \hline
10 & 399 & 410 & {\em ta} & B & 396 & 419 & 396 & 419 & 401 & 430 \\ \hline
11 & 441 & 473 & {\em tei yum} & B-HB & 441 & 473 & 441 & 473 & 443 & 471 \\ \hline
12 & 481 & 510 & {\em tat tat} & B-HB & 479 & 498 & 479 & 498 & 484 & 512 \\ \hline
13 & & & & & 502 & 513 & 502 & 513 & & \\ \hline
14 & 522 & 555 & {\em tei yum} & B-HB & 522 & 556 & 522 & 556 & 526 & 556 \\ \hline
15 & 564 & 574 & {\em ta} & B & 562 & 585 & 562 & 585 & 567 & 586 \\ \hline
16 & 605 & 638 & {\em tei yum} & B-HB & 607 & 639 & 607 & 639 & 608 & 638 \\ \hline
17 & 645 & 674 & {\em tat tat} & B-HB & 650 & 667 & 650 & 667 & 648 & 668 \\ \hline
18 & 686 & 719 & {\em tei yum} & B-HB & 688 & 717 & 688 & 717 & 689 & 716 \\ \hline
19 & 728 & 739 & {\em ta} & B & 724 & 773 & 724 & 773 & 730 & 760  \\ \hline
\end{tabular}
\end{scriptsize}
\end{table*}


\begin{table*}[!ht]
\caption{Results of Extraction of Key Frames in {\em Natta 8 Adavu}\label{tab:kp_natta8}}
\centering
\begin{scriptsize}
\begin{tabular}{|r||r|r|l|l||r|r||r|r||r|r|} \hline
\multicolumn{1}{|c||}{\bf Line} & \multicolumn{4}{c||}{\bf Audio Beats Mapped to} & \multicolumn{2}{c||}{\bf {\em No Motion}} & \multicolumn{2}{c||}{\bf Overlap of Beat} & \multicolumn{2}{c|}{\bf Annotation of} \\ 
\multicolumn{1}{|c||}{\bf \#} & \multicolumn{4}{c||}{\bf Video Frames} & \multicolumn{2}{c||}{\bf from Video} & \multicolumn{2}{c||}{\bf \& {\em No Motion}} & \multicolumn{2}{c|}{\bf {\em Key Posture}s} \\ \cline{2-11} 
\multicolumn{1}{|c||}{} & \multicolumn{1}{c|}{\bf Start} & \multicolumn{1}{c|}{\bf End} & \multicolumn{1}{c|}{\bf {\em Bol}} & \multicolumn{1}{c||}{\bf Beat} & \multicolumn{1}{c|}{\bf Start} & \multicolumn{1}{c||}{\bf End} & \multicolumn{1}{c|}{\bf Start} & \multicolumn{1}{c||}{\bf End} & \multicolumn{1}{c|}{\bf Start} & \multicolumn{1}{c|}{\bf End} \\ 
\multicolumn{1}{|c||}{} & \multicolumn{1}{c|}{\bf Frame} & \multicolumn{1}{c|}{\bf Frame} & \multicolumn{1}{c|}{\bf } & \multicolumn{1}{c||}{\bf Info} & \multicolumn{1}{c|}{\bf Frame} & \multicolumn{1}{c||}{\bf Frame} & \multicolumn{1}{c|}{\bf Frame} & \multicolumn{1}{c||}{\bf Frame} & \multicolumn{1}{c|}{\bf Frame} & \multicolumn{1}{c|}{\bf Frame} \\ \hline \hline
1 & & & & & 1 & 50 & & & & \\ \hline
2 & 63 & 74 & {\em tei yum} & B-HB & & & & & & \\ \hline
3 & & & & & 89 & 99 & & & 85 & 104 \\ \hline
4 & 108 & 138 & {\em tat tat} & B-HB & & & & & 116 & 125 \\ \hline
5 & 151 & 185 & {\em tei yum} & B-HB & & & & & & \\ \hline
6 & 195 & 207 & {\em ta} & B & 195 & 222 & 195 & 222 & 202 & 224 \\ \hline
7 & 238 & 271 & {\em tei yum} & B-HB & 246 & 285 & 246 & 285 & 246 & 270 \\ \hline
8 & 279 & 309 & {\em tat tat} & B-HB & 308 & 320 & 308 & 320 & 306 & 317 \\ \hline
9 & 322 & 355 & {\em tei yum} & B-HB & & & & & & \\ \hline
10 & 363 & 375 & {\em ta} & B & 373 & 389 & 373 & 389 & 371 & 389 \\ \hline
11 & 406 & 438 & {\em tei yum} & B-HB & & & & & 432 & 446 \\ \hline
12 & 446 & 475 & {\em tat tat} & B-HB & 461 & 471 & 461 & 471 & 453 & 466 \\ \hline
13 & 487 & 520 & {\em tei yum} & B-HB & & & & & & \\ \hline
14 & 528 & 539 & {\em ta} & B & 529 & 549 & 529 & 549 & 537 & 552 \\ \hline
15 & 570 & 603 & {\em tei yum} & B-HB & 576 & 609 & 576 & 609 & 574 & 604 \\ \hline
16 & 610 & 639 & {\em tat tat} & B-HB & 619 & 631 & 619 & 631 & 636 & 646 \\ \hline
17 & 651 & 684 & {\em tei yum} & B-HB & & & & & & \\ \hline
18 & 693 & 704 & {\em ta} & B & 697 & 737 & 697 & 737 & 702 & 732 \\ \hline
\end{tabular}
\end{scriptsize}
\end{table*}

\subsection{Results of Extraction of Key Frames}
We present the results of extraction of {\em K-frame}s for {\em Natta 3 Adavu} (Table~\ref{tab:kp_natta3}) and {\em Natta 8 Adavu} (Table~\ref{tab:kp_natta8}). In the tables the columns (a)--(b) present the sequence of audio event as in predicates~(\ref{eqn:audio_event}) (with annotations in columns (c)--(d)), columns (e)--(f) present the sequence of associated video  events as in predicates~(\ref{eqn:video_event}), and columns (g)--(h) present the sequence from predicates~(\ref{eqn:k_frames}). Finally, columns (i)--(j) present the ground truth from {\em K-frame} annotation (as in predicates~(\ref{eqn:annotation})) to be used for validation (predicates~(\ref{eqn:validate})).



In Table~\ref{tab:kp_natta3} ({\em Natta 3 Adavu}), no audio event exists for Line \#1 while there is a long range of {\em no motion} (about 3 sec. over 90 frames). This is when the dancer is in a stance and is just getting ready to start. In Line \#s 3 and 4, a single audio event has overlap with two separate {\em no motion} events that have very little gap (only one frame \# 162, which is due to noise, separates them). Consequently, both of these are detected as ranges of {\em K-frame}s. These should have got merged. Similar behavior can be observed in Line \#s 12 and 13 where 3 frames separate the ranges. Barring these two aberrations, the results match the annotations accurately.

On the other hand, we find a quite few mismatches in Table~\ref{tab:kp_natta8} ({\em Natta 8 Adavu}). This is a tricky case because this {\em Adavu} indeed does not have {\em K-frame}s to go with the 3$^{rd}$ beat (Line \#5), 7$^{th}$ beat (Line \#9), 11$^{th}$ beat (Line \#13) and 15$^{th}$ beat (Line \#17) (as can be seen from the annotations). There is trajectorial motion at these beats. Further, in Line \#3 {\em Key Posture} occurs quite some time after the audio beat. On recheck, we find this to be an artifact in the data itself. The {\em no motion} detection algorithm also fails to detect {\em Key Posture}s in two cases in Lines \#4 and \#11. Again, on scrutiny we find that the dancer actually is not momentarily stationary during this period; rather she is slowly moving over the entire range (possibly building up for or recovering from the motions mentioned above). Hence, {\em no motion} is not detected. 

In a similar manner, we validate the {\em K-frame}s as detected for the remaining {\em Adavu}s against the annotation. {\bf We achieve 83.49\% accuracy overall}.

\section{Recognition of {\em Key Posture}s}\label{sec:posture_recognizer}
Our next target is to recognize the {\em Key Posture}s in the extracted {\em K-frame}s. During the vocab generation process, we identify 23 unique {\em Key Posture}s (Figure~\ref{fig:natta_adavus_postures}) that occur in 8 {\em Natta Adavu}s.

%
%
%
%

During extraction of {\em K-frame} one can identify a range of RGB (and corresponding depth and skeleton) frames, all having {\em no motion}. Hence, these can be treated as equivalent and we can use any one of them for recognizing the {\em Key Posture} in them. This observation is utilized as follows: 
\begin{itemize}
\item Additional data sets may be generated by dividing a range of {\em K-frame}s in multiple parts (Section~\ref{sec:adavu_data_sets}).
\item While RGB frames are almost identical, the corresponding depth and skeleton frames, however, may not be identical due to IR noise and estimation error. Hence a sub-range of depth and / or skeleton frames may be used for better estimation of features (Section~\ref{sec:skeleton_features}).
\end{itemize}

\subsection{Features of Postures}
We use both skeleton and RGB streams for posture recognition. Angle features from skeleton and HOG features from RGB are extracted for this purpose. 

\subsubsection{Skeleton Features}\label{sec:skeleton_features}
\begin{figure}[!ht]
\centering
\begin{tabular}{c}
\includegraphics[width=5cm]{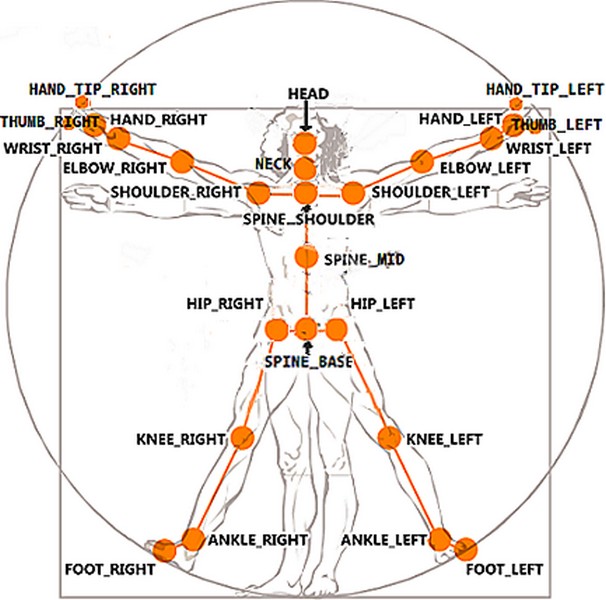} \\
{\footnotesize {\bf Source}: \url{https://msdn.microsoft.com/en-us/library/jj131025.aspx}}
\end{tabular}
\caption{Kinect Skeleton Joints\label{fig:skeleton}}
\end{figure}

Each skeleton frame (corresponding to an RGB frame) in a skeleton stream has 20 joint points (Figure~\ref{fig:skeleton}) given in terms of their 3D coordinates. Hence typical features of a skeleton frame include -- (1) the 3D coordinates of the joint points, (2) instantaneous velocity of the joint points, (3) instantaneous acceleration of the joint points, (4) lengths of the segments (bones) joining the joint points, (5) orientation of the segments joining the joint points, and so on. Most of these features, with the exception of the {\em angular orientation of the bones}, are not invariant to the scale, physical dimensions of the dancer, and minor variations in the postures. Hence we choose to use the angular orientation of the bones as descriptors of postures. 

The postures in a dance sequence are dominated by the positions of the legs and the arms. Hence, we choose to restrict our focus only to the 8 bones -- 4 each from the left and right sides of the body (Figure~\ref{fig:skeleton}):

\begin{itemize}
\item Bone joining Shoulder-Center (SPINE\_SHOULDER) and Shoulder: 
\begin{itemize}
\item SPINE\_SHOULDER -- SHOULDER\_LEFT 
\item SPINE\_SHOULDER -- SHOULDER\_RIGHT
\end{itemize}
\item Bone joining Shoulder and Elbow: 
\begin{itemize}
\item SHOULDER\_LEFT -- ELBOW\_LEFT
\item SHOULDER\_RIGHT -- ELBOW\_RIGHT
\end{itemize} 
\item Bone joining Hip-Center (SPINE\_BASE) and Hip: 
\begin{itemize}
\item SPINE\_BASE -- HIP\_LEFT  
\item SPINE\_BASE -- HIP\_RIGHT
\end{itemize}
\item Bone joining Hip and Knee: 
\begin{itemize}
\item HIP\_LEFT -- KNEE\_LEFT 
\item HIP\_RIGHT -- KNEE\_RIGHT
\end{itemize}
\end{itemize}

The angular orientation of every bone comprises 3 angles with X--, Y--, and Z--axes. Hence with 8 bones the feature vector would be 8 $\times$ 3 = 24 dimensional.

{\em In some frames, the skeletons are ill-formed due to IR noise and error in the skeleton estimation algorithm of Kinect. To work around this, we consider 5 consecutive skeleton frames from the range of {\em K-frame}s and take the average of the coordinates of the joint points. The angular orientation values are computed based on these average values. Also, we do not use the hand information as fingers are often not clear in the skeleton.}
 
\subsubsection{RGB Features}\label{sec:rgb_features}
Kinect skeletons suffer from variety of noise and may be ill-formed for a frame. Hence, we also use RGB features for posture recognition. First, we extract the human figure from the image using Kinect depth mask. Kinect processes the depth data to locate human body. It detects human body present in the imaging space and provides a binary mask of the human in the depth frame (in terms of markers for {\em players}). Using the mask we can extract the human from the depth image. But the depth and RGB images of Kinect are not aligned (because these are captured by two separate cameras on the device). Hence, we perform an affine transformation to bring the depth mask to the RGB space. We get the transformation matrix as a part of the data captured through {\em nuiCapture}. After getting the binary mask, we multiply the mask with the RGB image and convert the image into gray-scale. 

\begin{figure}[!ht]
\centering
\begin{tabular}{ccc}
\includegraphics[width=2.5cm]{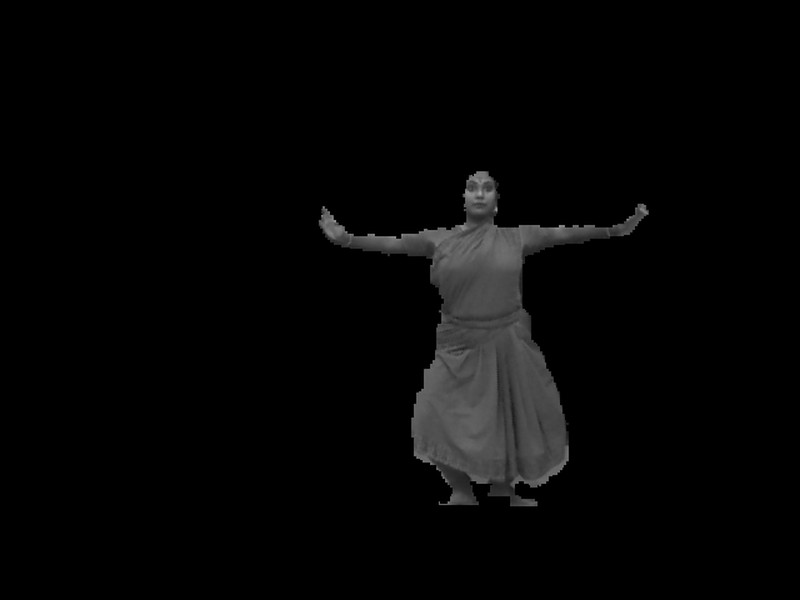} &
\includegraphics[width=2.5cm]{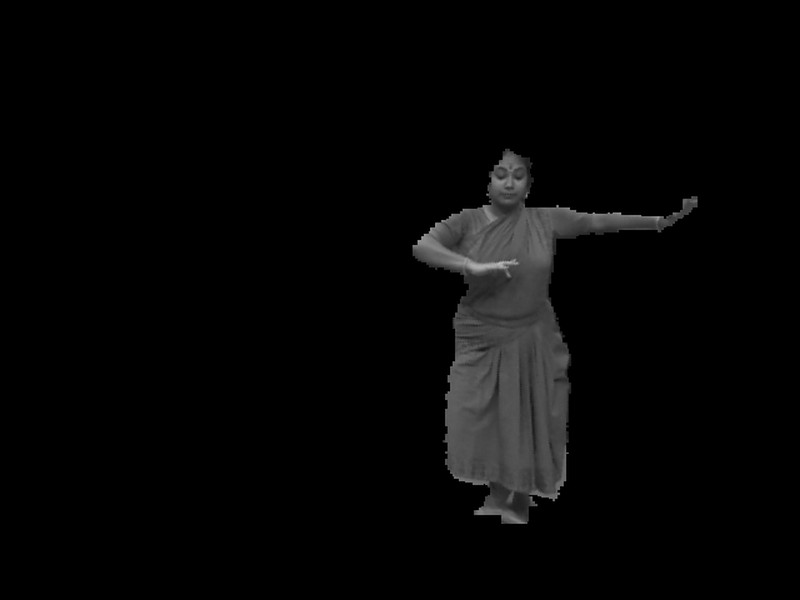} &
\includegraphics[width=2.5cm]{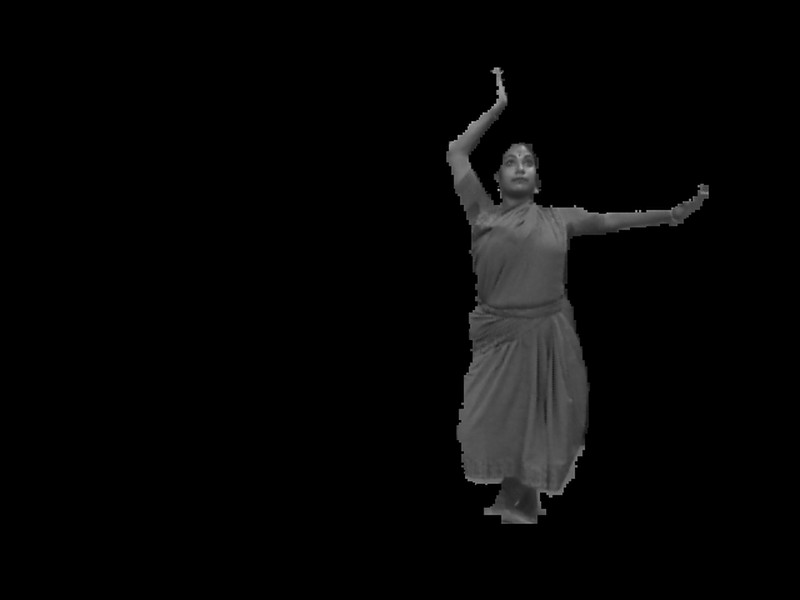} \\
\multicolumn{3}{c}{\small Results of Human Extraction} \\ \\
\includegraphics[width=2.5cm]{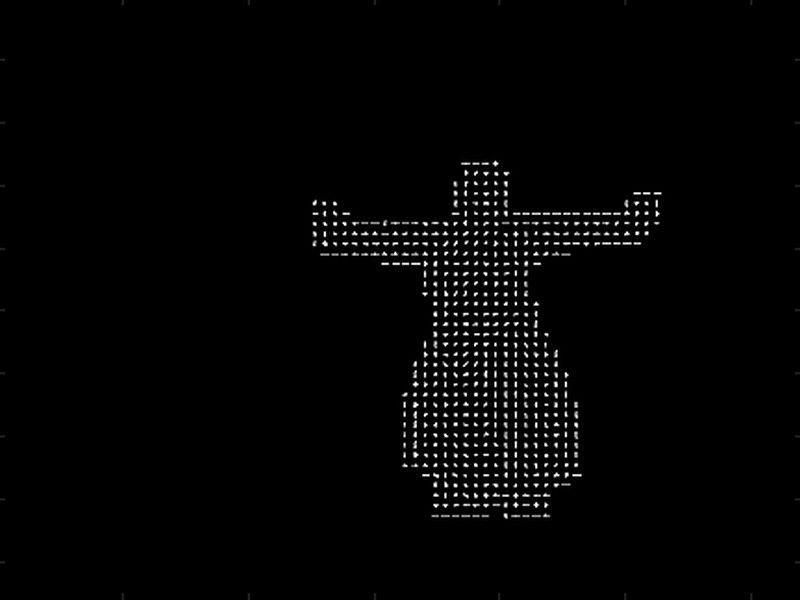} &
\includegraphics[width=2.5cm]{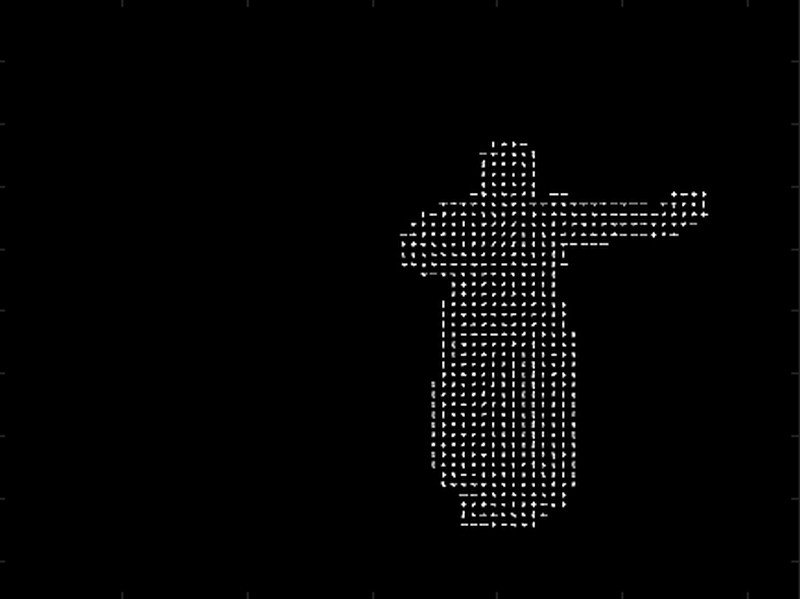} &
\includegraphics[width=2.5cm]{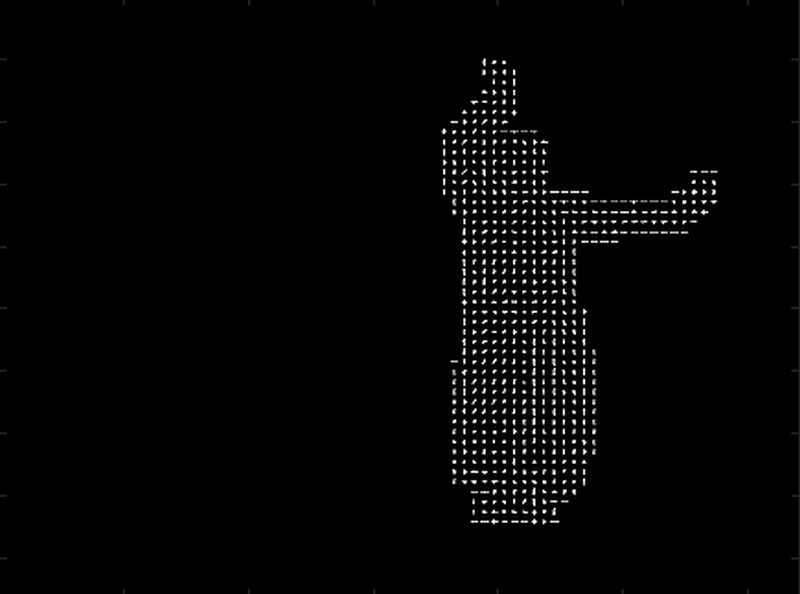} \\
\multicolumn{3}{c}{\small Visualization of HOG features} \\
\end{tabular}
\caption{RGB Features\label{tab:rgb_feature}}
\end{figure}

We next compute the {\em Histograms of Oriented Gradient} (HOG)~\cite{dalal2005histograms} descriptors for each posture frame. HOG is widely used for human detection in an image. This technique counts occurrences of gradient orientation in localized portions of an image. We use the HOG descriptor for recognition of the postures. 
HOG is calculated by dividing the input image into small spatial regions called {\em cell}s. 1-D histogram of gradient directions or edge orientations is calculated over each pixel of the cells. The energy of the local histograms are then normalized over a large spatial regions called {\em block}s to get the intended HOG feature descriptor. 
HOG feature length, $N$, depend upon on the size of the input image and the function parameter values such as cell size, block size, number of orientation bins used for the histogram and the number of overlapping cells between adjacent blocks. 
$N$ is equal to element wise product of $BlocksPerImage, BlockSize$, and $NumBins$, where $$BlocksPerImage = \frac{\frac{ImageSize}{CellSize} - BlockSize}{(BlockSize - BlockOverlap) + 1},$$ 
$ImageSize$ is the size of the input image, $CellSize$ is the size of the cell, $BlockSize$ is the number of cells in block, $NumBins$ is the number of orientation histogram bins, and $BlockOverlap$ is the number of overlapping cells between adjacent blocks. Considering 9-bin histogram of gradients over $8\times 8$ pixels sized cells, and blocks of $2\times 2$ cells, we extract 9576 dimensional HOG feature vector for each image of size 120 * 160. Sample postures and the plot of the corresponding HOG features are shown in Figure~\ref{tab:rgb_feature}.

\subsection{Recognition of {\em Key Posture}s -- Multiple Strategies}
Every dance is a sequence of different postures (not necessarily distinct) where each posture spans over a time range (sequence of {\em K-frame}s identified earlier). We get the various frames from Kinect corresponding to the time range of a posture. 

Recognizing the posture represented by these frames is a challenging task. Identifying the posture can help in many applications. For instance, once the posture has been identified, we can provide feedback to the dancer whether she / he is doing that posture correctly or not. 

To classify the postures, we use three classifiers (Table~\ref{tab:pos_result}): {\em GMM}, {\em SVM}, and {\em CNN}.


\subsubsection{Posture Recognition using GMM}\label{sec:gmm_recog}
We choose GMM~\cite{reynolds2015gaussian} as a classifier for posture recognition for the following reasons: 
\begin{itemize}
\item We expect the structures of the classes to follow normal distribution with different mean and variance. Hence a mixture model is a first choice.
\item GMM is the fastest algorithm for learning mixture models.
\item As GMM algorithm maximizes only the likelihood, it does not bias the means towards zero, or bias the cluster sizes to have specific structures that may or may not hold.
\item GMM comes with different options to constrain the covariance of the difference classes estimated: {\em spherical}, {\em diagonal}, {\em tied} or {\em full} covariance.
\end{itemize}
However, there are challenges too in using GMM:
\begin{itemize}
\item In GMM, each posture ({\em K-frame}) needs to be represented in the form of a feature vector. So feature selection is an important step. We decide to use skeleton features (Section~\ref{sec:skeleton_features}) for this.
\item The skeletal data from Kinect is noisy at times. It can affect the training phase of our classifier.
\item When there are inadequate observations per mixture, estimating the covariance matrices gets difficult, and the algorithm is known to diverge.
\end{itemize}

\subsubsection*{Parameter selection}
We use the scikit-learn library\footnote{\url{http://scikit-learn.org/stable/}} for implementation of GMM because it allows to learn the parameters of {\em Gaussian Mixture Model}s ({\em spherical}, {\em diagonal}, {\em tied} or {\em full} covariance matrices are supported) from data. It also facilitates to determine the appropriate number of components. We initially performed a few trials to understand the nature of the data and came up with the following parameters:

\begin{itemize}
\item {\bf Covariance matrix type}: With our experimental trials we achieve 84.7\%, 83.9\%, 98.4\%, and 97.7\% accuracy for {\em spherical}, {\em diagonal}, {\em full}, and {\em tied} covariance matrix types respectively. Hence, we choose `{\em full}' covariance matrix type for our model.

\item {\bf Number of iterations}: We need to decide on a value that neither under-fits nor over-fits our model on the training data set. Using the `{\em full}' covariance matrix -- we train our model multiple times using different number of iterations ($10^4$, $2\times10^4$, $5\times10^4$, $10^5$, $2\times10^5$, $5\times10^5$). We get the peak at $10^5$ and decide to use it for training our model. 
\end{itemize}



\subsubsection*{Results and Analysis}
We train and test our model using the data set in Table~\ref{tab:posture_data}. {\bf We achieve 83.04\% accuracy for posture recognition using GMM}. The confusion matrix is shown in Table~\ref{tab:conf_gmm} with major mis-classifications. The primary reasons of mis-classifications are:

\begin{table}[!ht]
\caption{Confusion matrix for Posture Recognition by GMM\label{tab:conf_gmm}}
\setlength{\tabcolsep}{2.5pt}
\centering 
\begin{scriptsize}
\begin{tabular}{l|c|r|p{1.2cm}|r||c|r|p{1.2cm}|r|} 
\multicolumn{1}{l}{\multirow{40}{*}{\textbf{\begin{rotate}{+90}{Actual Class \quad\quad\quad\quad}\end{rotate}}}}
& \multicolumn{8}{c}{\bf Predicted Class} \\
\multicolumn{1}{l}{ } &
\multicolumn{1}{c}{\bf Class} & 
\multicolumn{1}{c}{\textbf{Self}} &
\multicolumn{1}{c}{\textbf{Error}} &
\multicolumn{1}{c}{\textbf{Total}} &
\multicolumn{1}{c}{\bf Class} & 
\multicolumn{1}{c}{\textbf{Self}} &
\multicolumn{1}{c}{\textbf{Error}} &
\multicolumn{1}{c}{\textbf{Total}} \\ \cline{2-9}
& {\bf C01} & {\bf 96.0} &  &	1457 & {\bf C13} & {\bf 68.7} & (\alert{\bf 10.0, C15}), (\alert{\bf 6.2, C08}), (\alert{\bf 6.2, C11})  &	80 \\

& {\bf C02} & {\bf 87.2} & (\alert{\bf 5.7, C01}) &	873	
& {\bf C14} & {\bf 67.5} & (\alert{\bf 31.6, C01}) & 117 \\

& {\bf C03} & {\bf 85.9} & &	561	
& {\bf C15} & {\bf 66.9} & (\alert{\bf 9.0, C01}) &	121 \\

& {\bf C04} & {\bf 62.5} & (\alert{\bf 7.3, C10}), (\alert{\bf 6.8, C02}), (\alert{\bf 5.9, C01})  &	219	
& {\bf C16} & {\bf 33.3} & (\alert{\bf 20.8, C07}), (\alert{\bf 18.7, C03}), (\alert{\bf 16.6, C02}) &	48 \\

& {\bf C05} & {\bf 75.3} & (\alert{\bf 5.2, C01}) &	268	
& {\bf C17} & {\bf 19.6} & (\alert{\bf 15.6, C01}), (\alert{\bf 15.6, C03}), (\alert{\bf 13.7, C06}), (\alert{\bf 9.8, C09}), (\alert{\bf 5.8, C02}), (\alert{\bf 5.8, C12}) &	51 \\

& {\bf C06} & {\bf 85.0} & (\alert{\bf 5.1, C01}) &	541	
& {\bf C18} & {\bf 61.7} &(\alert{\bf 14.8, C01}), (\alert{\bf 8.6, C02}), (\alert{\bf 6.1, C03}), (\alert{\bf 6.1, C05}) & 81 \\

& {\bf C07} & {\bf 86.1} & & 475	
& {\bf C19} & {\bf 76.0} & (\alert{\bf 10.8, C06}), (\alert{\bf 8.6, C04}) & 46 \\

& {\bf C08} & {\bf 63.3} & (\alert{\bf 18.7, C01}) & 112	
& {\bf C20} & {\bf 67.4} & (\alert{\bf 6.9, C05}), (\alert{\bf 6.9, C07}), (\alert{\bf 6.9, C12}) &	43 \\

& {\bf C09} & {\bf 75.1} & (\alert{\bf 14.2, C15}) 	&	133	
& {\bf C21} & {\bf 50.0} & (\alert{\bf 16.6, C09}, (\alert{\bf 16.6, C11}, (\alert{\bf 16.6, C13} &	6 \\

& {\bf C10} & {\bf 74.6} & (\alert{\bf 7.4, C02}), (\alert{\bf 6.7, C01}) & 162	
& {\bf C22} & {\bf 33.3} & (\alert{\bf 16.6, C05}), (\alert{\bf 16.6, C08}), (\alert{\bf 16.6, C11}), (\alert{\bf 16.6, C15}) & 6 \\

& {\bf C11} & {\bf 95.7} & &	117	
& {\bf C23} & {\bf 45.9} & (\alert{\bf 13.1, C03}), (\alert{\bf 11.4, C08}) & 61 \\

& {\bf C12} & {\bf 70.2} & (\alert{\bf 23.8, C02}), (\alert{\bf 5.9, C01}) 	& 84	
&		&		&		&	 \\ 
\cline{2-9}
\multicolumn{9}{c}{ } \\
\multicolumn{1}{c}{ } & \multicolumn{8}{p{7cm}}{For test data (Tab.~\ref{tab:posture_data}), the diagonal entries (in \%) of the confusion matrix are shown as `{\bf Self}'. Entries with 5\%+ error are shown under `{\bf Error}'. For example, for {\bf {\em Class}} = {\bf C14}, the diagonal entry is 67.5\% and it is mis-classified as {\bf C01} in 31.6\% cases. `{\bf Total}' shows the number of symbols in the class}  \\
\end{tabular}
\end{scriptsize}
\end{table}

\begin{itemize}
\item {\bf Similarity of Postures}: In Table~\ref{tab:conf_gmm} we find that posture {\bf C13} is classified as posture {\bf C15} in 10\% of the cases. If we look at the posture {\bf C13} and {\bf C15} in Figure~\ref{fig:natta_adavus_postures}, we observe that they are quite similar in the way the hands and legs are stretched. Hence strong posture similarity causes mis-classification.

\item {\bf Skeleton Noise}: At times two postures that are quite different from each other may still have quite similar skeletons. This is due to IR noise and error in the skeleton estimation algorithm of Kinect. The error may span several consecutive frames so that our strategy of averaging over 5 consecutive skeletal frames fail. For example, in Figure~\ref{fig:skel_noise_c14_c01} the right leg of posture {\bf C14} is not stretched at all in the skeleton image. Hence, the skeleton of {\bf C14} looks like the skeleton of {\bf C01}. On the other hand, the noise is very high in the skeletons of {\bf C16} and {\bf C17} (Figure~\ref{fig:skel_noise_c16_c17}). Hence {\bf C16} and {\bf C17} are mis-classified with several other classes. 

\begin{figure}[!ht]
\centering
\begin{small}
\begin{tabular}{cccc}
\includegraphics[width=3cm]{Images/2nP14} &
\includegraphics[width=3cm]{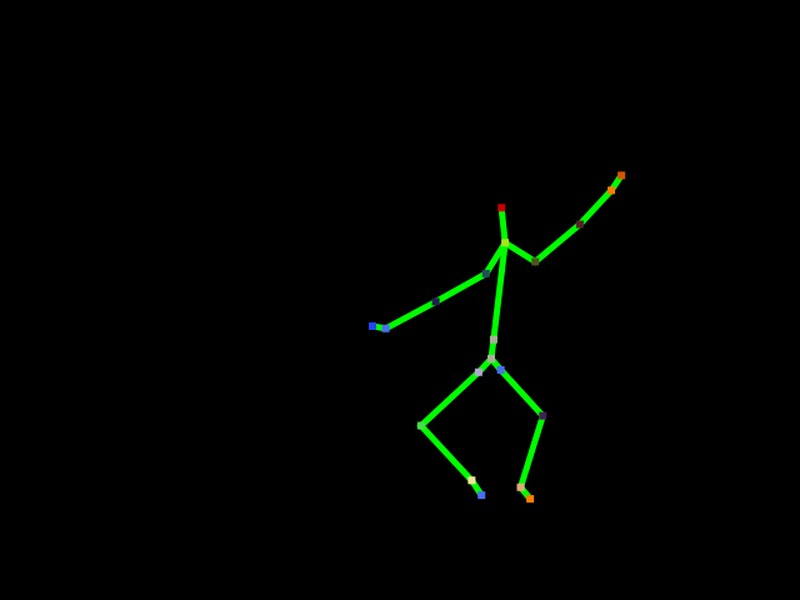}&
\includegraphics[width=3cm]{Images/2nP01} &
\includegraphics[width=3cm]{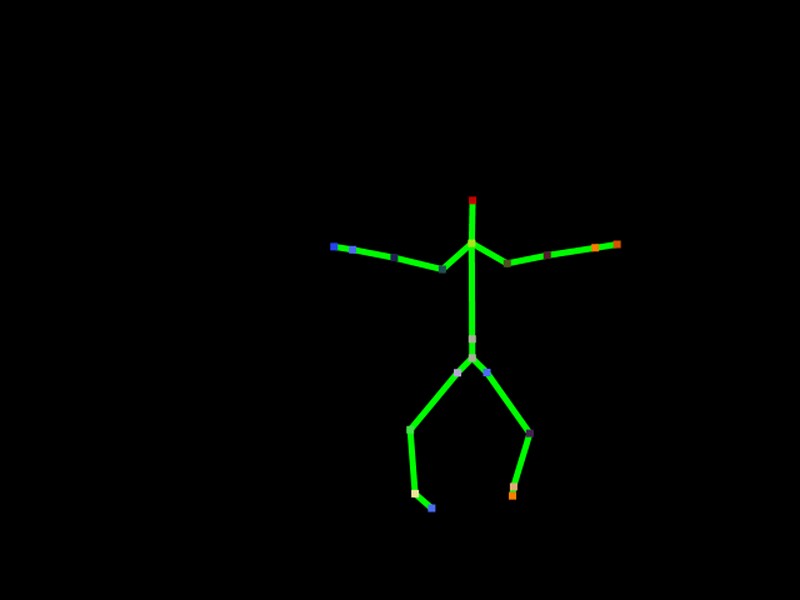} \\
{\bf C14} & {\bf C01} &  Skeleton of {\bf C14} & Skeleton of {\bf C01}\\
\end{tabular}
\end{small}
\caption{Skeleton noise leading to confusion between {\bf C14} and {\bf C01}\label{fig:skel_noise_c14_c01}}
\end{figure}

\begin{figure}[!ht]
\centering
\begin{small}
\begin{tabular}{cccc}
\includegraphics[width=3cm]{Images/2nP16} &
\includegraphics[width=3cm]{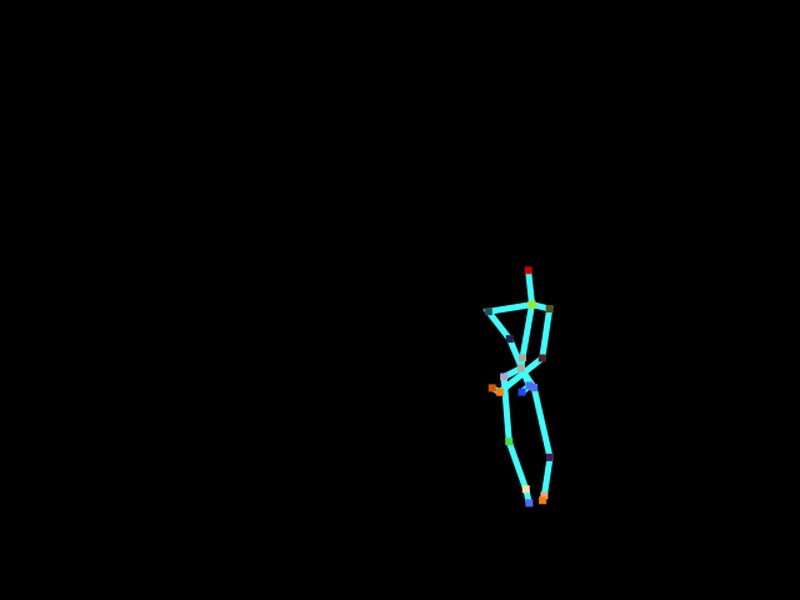} &
\includegraphics[width=3cm]{Images/2nP17} &
\includegraphics[width=3cm]{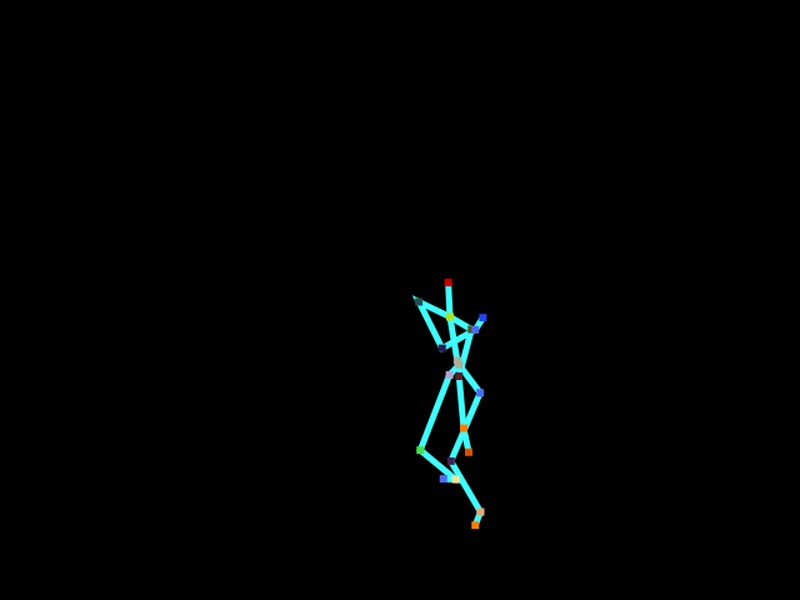} \\
{\bf C16} & {\bf C17} & Skeleton of {\bf C16} & Skeleton of {\bf C17} \\

\end{tabular}
\end{small}
\caption{Skeleton noise and Symmetry in {\bf C16} and {\bf C17}\label{fig:skel_noise_c16_c17}}
\end{figure}


\end{itemize}

\subsubsection{Posture Recognition using SVM}\label{sec:svm_recog}
Skeleton noise adversely impacts the accuracy of posture recognition by GMM using the angular orientation features from skeleton images. Hence, we next explore recognition by {\em Support Vector Machine} (SVM) using HOG features from RGB images. SVM (\cite{vapnik2013nature, scholkopf1997comparing, suykens1999least}) is one of the most widely used supervised learning algorithms. The advantages of SVM as a classifier include:
\begin{itemize}
\item SVM is effective where the number of dimensions is greater than the number of samples.
\item The support vectors use subset of the training points. Hence the classifier is often efficient.
\item Different Kernel functions can be used as the decision function based on the training data set.
\end{itemize}

The challenges in using SVM include:
\begin{itemize}
\item The classifier may give poor performance if the number of features is much greater than the number of samples.
\item The classifier does not provide probability estimates.
\end{itemize}

SVM is a binary classifier, that is, it separates only between 2 classes. However multiple classes can be classified by using multiple SVMs. SVM is a maximal margin classifier, which finds a hyper-plane (decision boundary) which separates the 2 classes by maximizing the distance of the margin from the 2 classes.

\subsubsection*{Parameter Selection}
Optimal choice of kernel functions for an SVM is an open problem in the literature. There is no known method to guarantee that a particular kernel function will consistently perform better than others. Hence the choice of kernel functions varies on a case to case basis. Two of the commonly used kernels are {\em Linear Kernel} and {\em Gaussian Kernel}. If the data is non-linear, Gaussian kernel is usually the best for many applications. This kernel is also known as {\em Radial Basis Function} (RBF) kernel and is defined as:
$$K(x, x'i) = \exp\left \{ -\frac{\|x-x'\|^2}{2\sigma^2}\right \}$$
where $x$ AND $x'$ are two samples represented as feature vectors in some input space, $\|x-x'\|^2$ is the squared Euclidean distance between the two feature vectors, and $\sigma$ is a free parameter.
\subsubsection{Results and Analysis}
To classify the postures into 23 posture classes (Figure~\ref{fig:natta_adavus_postures}), we use {\em One vs. Rest} type of multi-class SVM. Hence, only 23 models is trained as opposed to $\Comb{23}{2}$ = 253 models that would have been necessary for {\em One vs. One} type. The data set from Table~\ref{tab:posture_data} is used for training and testing the SVM. For testing we use the trained SVM models to predict the class labels. {\bf We achieve 97.95\% accuracy for posture recognition using SVM}. 

The confusion matrix is shown in Table~\ref{tab:conf_svm}. The table shows that the major mis-classifications occur between {\bf C16}--{\bf C17} and {\bf C19}--{\bf C20}. The posture classes {\bf C16}--{\bf C17} (Figure~\ref{fig:skel_noise_c16_c17}) and {\bf C19}--{\bf C20} (Figure~\ref{fig:natta_adavus_postures}) are actually symmetric and strongly similar. Therefore, one class gets mis-classified to other symmetric class. 


\begin{table}[!ht]
\caption{Confusion matrix for Posture Recognition by SVM\label{tab:conf_svm}}
\setlength{\tabcolsep}{2.5pt}
\centering 
\begin{scriptsize}
\begin{tabular}{l|c|r|p{1.2cm}|r||c|r|p{1.2cm}|r|} 
\multicolumn{1}{l}{\multirow{23}{*}{\textbf{\begin{rotate}{+90}{Actual Class \quad\quad\quad\quad}\end{rotate}}}}
& \multicolumn{8}{c}{\bf Predicted Class} \\
\multicolumn{1}{l}{ } &\multicolumn{1}{c}{\bf Class} & 
\multicolumn{1}{c}{\textbf{Self}} &
\multicolumn{1}{c}{\textbf{Error}} &
\multicolumn{1}{c}{\textbf{Total}} 
&\multicolumn{1}{c}{\bf Class} & 
\multicolumn{1}{c}{\textbf{Self}} &
\multicolumn{1}{c}{\textbf{Error}} &
\multicolumn{1}{c}{\textbf{Total}} \\ \cline{2-9}
&	 {\bf C01} 	&	 {\bf 100.0} 	&	 	&	1457	&	 {\bf C12} 	&	 {\bf 100.0} 	&	 	&	84 \\
&	 {\bf C02} 	&	 {\bf 99.8} 	&	 (\alert{\bf 0.2, C01}) 	&	873	&	 {\bf C13} 	&	 {\bf 100.0} 	&	 	&	 80 \\
&	 {\bf C03} 	&	 {\bf 100.0} 	&	 	&	561	&	 {\bf C14} 	&	 {\bf 100.0} 	&	 	&	 117 \\
&	 {\bf C04} 	&	 {\bf 100.0} 	&	 	&	219	&	 {\bf C15} 	&	 {\bf 100.0} 	&	 	&	 121 \\
&	 {\bf C05} 	&	 {\bf 100.0} 	&	 	&	268	&	 {\bf C16} 	&	 {\bf 100.0} 	&	 	&	 48 \\
&	 {\bf C06} 	&	 {\bf 100.0} 	&	 	&	541	&	 {\bf C17} 	&	 {\bf 49.0}  	&	 (\alert{\bf 51.0, C16}) 	&	 51 \\
&	 {\bf C07} 	&	 {\bf 100.0} 	&	 	&	475	&	 {\bf C18} 	&	 {\bf 100.0} 	&	 	&	 81 \\
&	 {\bf C08} 	&	 {\bf 100.0} 	&	 	&	112	&	 {\bf C19} 	&	 {\bf 39.1}  	&	 (\alert{\bf 23.9, C01}), (\alert{\bf 17.4, C04}), (\alert{\bf 19.6, C20}) 	&	 46 \\
&	 {\bf C09} 	&	 {\bf 100.0} 	&	 	&	133	&	 {\bf C20} 	&	 {\bf 55.8} 	&	 (\alert{\bf 44.2, C19}) 	&	 43 \\
&	 {\bf C10} 	&	 {\bf 100.0} 	&	 	&	162	&	 {\bf C21} 	&	 {\bf 100.0} 	&	 	&	 6 \\
&	 {\bf C11} 	&	 {\bf 65.0}  	&	 (\alert{\bf 15.4, C01}), (\alert{\bf 19.6, C05}) 	&	117	&	 {\bf C22} 	&	 {\bf 100.0} 	&	 	&	 6 \\
&		&		&		&		&	 {\bf C23} 	&	 {\bf 100.0} 	&	 	&	 61 \\ \cline{2-9}
\multicolumn{9}{c}{ } \\
\multicolumn{1}{c}{ } & \multicolumn{8}{p{7cm}}{For test data (Tab.~\ref{tab:posture_data}), the diagonal entries (in \%) of the confusion matrix are shown as `{\bf Self}'. Entries with non-zero error are shown under `{\bf Error}'. For example, for {\bf {\em Class}} = {\bf C17}, the diagonal entry is 49.0\% and it is mis-classified as {\bf C16} in 51.0\% cases. `{\bf Total}' shows the number of symbols in the class}  \\
\end{tabular}
\end{scriptsize}
\end{table}

\subsubsection{Posture Recognition using CNN}\label{sec:cnn_recog}
We observe that GMM and SVM classifiers for posture recognition are able to perform well on crafted features like skeleton angle and HOG. But the traditional feature engineering has a few challenges:
\begin{itemize}
\item There are limitations in abstracting various higher levels of body features.
\item It is highly challenging to model natural variations that exists in the posture structures.
\end{itemize}
With recent advancement in GPU technology, the deep learning technique is able to achieve state-of-the-art recognition in many image classification problems if fairly large training data set is available (as in our case). Moreover, with deep learning, we do not need to manually select the features. Hence we explore {\em Convolutional Neural Network} (CNN) to deal with the posture recognition problem.
The challenges in using CNN include:
\begin{itemize}
\item The data set size for each dance posture is uneven. Few postures are more likely to occur and are more in number. But few postures which do not occur so frequently are less in number. For example (Table~\ref{tab:posture_data}), posture {\bf C01} (Figure~\ref{fig:natta_adavus_postures}) occurs frequently while a difficult posture like {\bf C22} occurs in hardly few {\em Adavu}s, leading to small data size.


\item We need to decide on a proper architecture, that is, how many layers should be there and how many nodes should exist in each layer. This decision may require several trials.
\end{itemize}

\subsubsection{Dataset}\label{sec:dataset}
Using the data set of Table~\ref{tab:posture_data} we {\bf get a test accuracy of 61.30\% for 23 classes}. The poor accuracy is due to very less data in some classes and symmetric postures between a number of classes.

The classes {\bf C21} and {\bf C22} (Table~\ref{tab:posture_data}) have hardly any data compared to class {\bf C01}. To handle this disparity we discard a few classes which have insufficient data. Hence, we ignore classes {\bf C12}, {\bf C13}, {\bf C18}, {\bf C21}, {\bf C22}, and {\bf C23}. And we merge classes {\bf C16}--{\bf C17} (Figure~\ref{fig:skel_noise_c16_c17}) and {\bf C19}--{\bf C20} (Figure~\ref{fig:natta_adavus_postures}) that corresponded to symmetric postures. Finally we have 15 classes (Table~\ref{tab:cnn_posture_data}). {\em Note that we renumber the classes after elimination and merging.}

%

\begin{table}[!ht]
\centering 
\caption{Restricted data set for CNN\label{tab:cnn_posture_data}}
\begin{tabular}{|l|r|r||l|r|r|} \hline
\multicolumn{1}{|c}{\bf Posture} & \multicolumn{1}{|c}{\bf Training} & \multicolumn{1}{|c}{\bf Test} & \multicolumn{1}{||c}{\bf Posture} & \multicolumn{1}{|c}{\bf Training} & \multicolumn{1}{|c|}{\bf Test} \\ 
\multicolumn{1}{|c}{\bf ID} & \multicolumn{1}{|c}{\bf data} & \multicolumn{1}{|c}{\bf data} & \multicolumn{1}{||c}{\bf ID} & \multicolumn{1}{|c}{\bf data} & \multicolumn{1}{|c|}{\bf data} \\ \hline \hline
 {\bf C01} & 6154 & 1457 & {\bf C09} & 306 & 133 \\
 {\bf C02} & 3337 & 873	& {\bf C10} & 397 & 162 \\
 {\bf C03} & 3279 & 561 & {\bf C11} & 408 & 117 \\
 {\bf C04} & 1214 & 219 & {\bf C12} & 393 & 117\\
 {\bf C05} & 1192 & 268 & {\bf C13} & 484 & 121 \\
 {\bf C06} & 1419 & 541 & {\bf C14} & 394 & 99 \\
 {\bf C07} & 1250 & 475 & {\bf C15} & 356 & 89 \\
 {\bf C08} & 284 & 112 & && \\ \hline
\end{tabular}
\end{table}


\subsubsection{Design of CNN}
\begin{figure}[!ht]
\centering
\includegraphics[width=8cm]{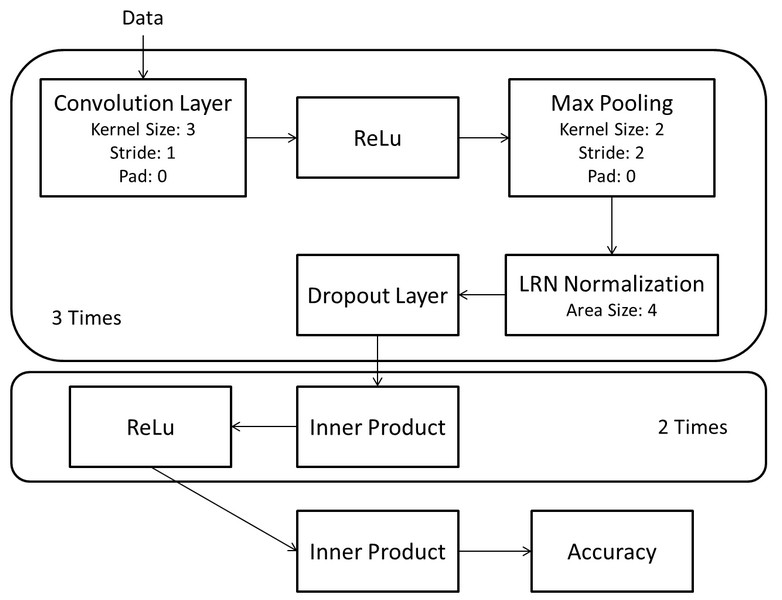}
\caption{CNN architecture for Posture Recognition\label{fig:AlexNet}}
\end{figure}

First we design each sub-layers of our network. The selected sub-layers are convolution, reLU, Max-Pooling, Local Response Normalization (LRN), Dropout layer and Inner product. We consider the standard way of organizing the layers, that is, convolution, relu and pooling few times and then finally ending with fully connected layers. We start with an initial architecture having a single convolution hidden layer and gradually increase it to obtain a better validation accuracy. With a number of trials, we obtain the best results with 3 hidden layers.
These convolution layers are then followed by two fully connected layers so that the final obtained result has dimension same as that of the number of classes. We also include a final loss layer which includes a Softmax followed by cross entropy. Figure~\ref{fig:AlexNet} shows the architecture.


\subsubsection{Results and Analysis}
We use the tensorflow library\footnote{\url{https://www.tensorflow.org/}} in Python language to implement the CNN. We train the CNN model on the training data set (Table~\ref{tab:cnn_posture_data}) and then test the model on a different unseen data set. {\bf We achieve accuracy of 99.12\% for posture recognition using CNN}. The confusion matrix (Tab.~\ref{tab:conf_cnn}) shows little significant error in classification.

\begin{table}[!ht]
\caption{Confusion matrix for Posture Recognition by CNN\label{tab:conf_cnn}}
\setlength{\tabcolsep}{2.5pt}
\centering 
\begin{scriptsize}
\begin{tabular}{l|c|r|p{1.1cm}|r||c|r|p{1.1cm}|r|} 
\multicolumn{1}{l}{\multirow{21}{*}{\textbf{\begin{rotate}{+90}{Actual Class \quad\quad\quad\quad}\end{rotate}}}}
& \multicolumn{8}{c}{\bf Predicted Class} \\
\multicolumn{1}{l}{ } &\multicolumn{1}{c}{\bf Class} & 
\multicolumn{1}{c}{\textbf{Self}} &
\multicolumn{1}{r}{\textbf{Error}} &
\multicolumn{1}{r}{\textbf{Total}} 
&\multicolumn{1}{c}{\bf Class} & 
\multicolumn{1}{c}{\textbf{Self}} &
\multicolumn{1}{r}{\textbf{Error}} &
\multicolumn{1}{r}{\textbf{Total}} \\ \cline{2-9}
&	 {\bf C01} 	&	 {\bf 100.0} 	&	 	&	1457	&	 {\bf C08} 	&	 {\bf 97.3} 	&	 (\alert{\bf 1.8, C01}), (\alert{\bf 0.9, C02}) 	&	 112 \\
&	 {\bf C02} 	&	 {\bf 99.7} 	&	 (\alert{\bf 0.2, C03}), (\alert{\bf 0.1, C01}) 	&	873	&	 {\bf C09} 	&	 {\bf 97.0} 	&	 (\alert{\bf 2.3, C01}), (\alert{\bf 0.8, C05}) 	&	 133 \\
&	 {\bf C03} 	&	 {\bf 99.3} 	&	 (\alert{\bf 0.7, C01}) 	&	561	&	 {\bf C10} 	&	 {\bf 99.4} 	&	 (\alert{\bf 0.6, C01}) 	&	 162 \\
&	 {\bf C04} 	&	 {\bf 98.6} 	&	 (\alert{\bf 1.4, C01}) 	&	219	&	 {\bf C11} 	&	 {\bf 97.4} 	&	 (\alert{\bf 1.7, C05}), (\alert{\bf 0.9, C01}) 	&	 117 \\
&	 {\bf C05} 	&	 {\bf 98.9} 	&	 (\alert{\bf 1.1, C01}) 	&	268	&	 {\bf C12} 	&	 {\bf 95.7} 	&	 (\alert{\bf 4.3, C01}) 	&	 117 \\
&	 {\bf C06} 	&	 {\bf 97.8} 	&	 (\alert{\bf 1.7, C01}), (\alert{\bf 0.4, C04}), (\alert{\bf 0.2, C02}) 	&	541	&	 {\bf C13} 	&	 {\bf 98.3} 	&	 (\alert{\bf 1.7, C01}) 	&	 121 \\
&	 {\bf C07} 	&	 {\bf 99.8} 	&	 (\alert{\bf 0.2, C06}) 	&	475	&	 {\bf C14} 	&	 {\bf 100.0} 	&	 	&	 99 \\
&		&		&		&		&	 {\bf C15} 	&	 {\bf 96.6} 	&	 (\alert{\bf 3.4, C01}) 	&	 89 \\  \cline{2-9}
\multicolumn{9}{c}{ } \\
\multicolumn{1}{c}{ } & \multicolumn{8}{p{7cm}}{Results for test data as in Table~\ref{tab:posture_data}. For test data (Tab.~\ref{tab:posture_data}), the diagonal entries (in \%) of the confusion matrix are shown as `{\bf Self}'. Entries with non-zero error are shown under `{\bf Error}'. For example, for {\bf {\em Class}} = {\bf C15}, the diagonal entry is 96.6\% and it is mis-classified as {\bf C01} in 3.4\% cases. `{\bf Total}' shows the number of symbols in the class}  \\
\end{tabular}
\end{scriptsize}
\end{table}

\subsection{Summary Results of {\em Key Posture} Recognition}
Three different classifiers have been designed with different input features to recognize {\em Key Posture}s. The accuracy results are summarized in Tab.~\ref{tab:pos_result}. We get the best recognition rate with CNN but using only 15 classes while SVM does commendably well for 23 classes. We observe that the traditional features give better results when the data set is skewed across classes -- specifically when some classes have very few number of samples. 

\begin{table}[!ht]
\centering
\caption{Results of Posture Recognition\label{tab:pos_result}}
\begin{small}
\begin{tabular}{|l|l|l|r|r|} \hline
\textbf{Recognizer} & \textbf{Input Data} & \textbf{Features} & \multicolumn{1}{l|}{\textbf{\begin{tabular}[c]{@{}c@{}}No. of \\ Classes\end{tabular}}} & \multicolumn{1}{l|}{\textbf{\begin{tabular}[c]{@{}c@{}}Recognition \\ Rate (\%)\end{tabular}}} \\ \hline\hline
{\em GMM} & Skeleton & Angle & 23	 & 83.04 \\ \hline 
{\em SVM} & RGB & HOG & 23	 & 97.95 \\ \hline 
{\em CNN} & RGB & - & 15	& 99.12 \\ \hline 
\end{tabular}
\end{small}
\end{table}

\section{{\em Adavu} Recognition\label{sec:adavu_recognizer}}
Using the extracted {\em K-frame}s and the {\em Key Posture}s as recognized above, we attempt to recognize 8 {\em Natta Adavu}s. That is, to recognize the {\em Adavu} from a sequence of {\em Key Posture}s corresponding to its performance. 
Identifying the dance as a particular {\em Adavu} is an interesting and challenging task. It can help in many applications. For instance, once the {\em Adavu} is identified, we can offer feedback to the dancer regarding his / her performance. The feedback can vary from being at a macro level like proper sequencing of steps to micro level like proper formation of a posture. Thus it can help build a dance tutoring system.

We have tried to recognize {\em Adavu}s or the posture sequences using HMM. To recognize the sequences we create one HMM for each {\em Adavu} class. Given a classifier of 8 {\em Adavu}s, we choose the model which best matches the observation from 8 HMMs, $\lambda_{HMM_i} = \{A_i, B_i, \pi_i \}, i = 1, ..., 8$. For an unknown sequence, we calculate $P(\lambda_{HMM_i}|O)$ for each HMM $\lambda_{HMM_i}$ and select $\lambda_c$ where $c = arg\; max \; P(\lambda_{HMM_i}|O)$. In the learning phase, each HMM is trained so that it is most likely to generate the symbol pattern of its category. Leaning of HMM means optimizing the model parameters ($A, B, \pi $) to maximize the probability of the observation sequence $P(O|\lambda_{HMM})$.

\subsection{Data Set}\label{sec:adavu_data_sets}
The data set for training and testing the HMMs is shown in Table~\ref{tab:adavu_data}. The natural pattern of posture transition in all 8 {\em Adavu}s are shown in Table~\ref{tab:sequence}. Here, the $\rightarrow$ indicate the {\em followedBy} relation between two posture classes.


\begin{table}[!ht]
\centering 
\caption{Sequence data set for {\em Adavu} Recognition\label{tab:adavu_data}}
\begin{tabular}{|r|r|r||r|r|r|} \hline
\multicolumn{1}{|c}{\bf {\em Adavu}} & \multicolumn{1}{|c}{\bf Training} & \multicolumn{1}{|c}{\bf Test} & \multicolumn{1}{||c}{\bf {\em Adavu}} & \multicolumn{1}{|c}{\bf Training} & \multicolumn{1}{|c|}{\bf Test} \\ 
\multicolumn{1}{|c}{\bf } & \multicolumn{1}{|c}{\bf data} & \multicolumn{1}{|c}{\bf data} & \multicolumn{1}{||c}{\bf } & \multicolumn{1}{|c}{\bf data} & \multicolumn{1}{|c|}{\bf data} \\ \hline \hline
{\em Natta 1} & 35 & 7 & {\em Natta 5} & 71 & 7 \\
{\em Natta 2} & 35 & 7 & {\em Natta 6} & 71 & 7 \\
{\em Natta 3} & 71 & 7 & {\em Natta 7} & 71 & 7 \\
{\em Natta 4} & 68 & 7 & {\em Natta 8} & 71 & 7 \\ \hline
\end{tabular}
\end{table}

\subsection{Training the HMM}
We build one HMM for each {\em Adavu}. The number of hidden states of the HMMs is considered to be equal to the number {\em Key Posture}s in an {\em Adavu}. A {\em K-frame} is represented in terms of the skeleton angles. {\em Estimation Maximization} (EM)~\cite{wilson1999parametric} algorithm is used for leaning the parameters of HMMs. The HMM with Gaussian emissions is employed. Diagonal covariance matrices are used for the Gaussian to exploit correlations between the elements of each observation. Additionally, the Viterbi algorithm~\cite{rabiner1989tutorial} is used for decoding. In total 8 HMMs are built to recognize 8 {\em Adavu} classes. To recognize a test {\em Adavu}, we compute the log probability under each model and select the model having  maximum log probability to predict the {\em Adavu}.

\subsection{Results of {\em Adavu} Recognition}
Using 8 variants of {\em Natta Adavu}, we {\bf achieve 94.64\% accuracy in recognition of {\em Natta Adavu}s} (Table~\ref{tab:seq}). The data set for {\em Adavu} recognition is given in Table~\ref{tab:adavu_data}. Mis-classification occurs between {\em Natta Adavu} classes 1 and 2 due to the  similarity between the natural patterns of these two sequences. 

\begin{table*}[!ht]
\caption{Sequence of postures in {\em Natta Adavu}s\label{tab:sequence}} 
\centering 
\begin{footnotesize}
\begin{tabular}{|c | l|} \hline
\multicolumn{1}{|c|}{\bf {\em Adavu}} & \multicolumn{1}{c|}{\bf Sequence of Postures Classes} \\ \hline \hline
{\em Beats (full)} &\textcolor{red}{B01$\rightarrow$B02$\rightarrow$B03$\rightarrow$B04$\rightarrow$B05$\rightarrow$B06$\rightarrow$B07$\rightarrow$B08$\rightarrow$}\textcolor{blue}{B09$\rightarrow$B10$\rightarrow$B11$\rightarrow$B12$\rightarrow$B13$\rightarrow$B14$\rightarrow$B15$\rightarrow$B16} \\ \hline\hline
{\em Natta 1} & {\bf C02}$\rightarrow${\bf C01}$\rightarrow${\bf C03}$\rightarrow${\bf C01}$\rightarrow${\bf C02}$\rightarrow${\bf C01}$\rightarrow${\bf C03}$\rightarrow${\bf C01}$\rightarrow${\bf C02}$\rightarrow${\bf C01}$\rightarrow${\bf C03}$\rightarrow${\bf C01}$\rightarrow${\bf C02}$\rightarrow${\bf C01}$\rightarrow${\bf C03}$\rightarrow${\bf C01} \\ \hline
{\em Natta 2} & {\bf C02}$\rightarrow${\bf C01}$\rightarrow${\bf C02}$\rightarrow${\bf C01}$\rightarrow${\bf C03}$\rightarrow${\bf C01}$\rightarrow${\bf C03}$\rightarrow${\bf C01}$\rightarrow${\bf C02}$\rightarrow${\bf C01}$\rightarrow${\bf C02}$\rightarrow${\bf C01}$\rightarrow${\bf C03}$\rightarrow${\bf C01}$\rightarrow${\bf C03}$\rightarrow${\bf C01} \\ \hline
{\em Natta 3} & {\bf C02}$\rightarrow${\bf C01}$\rightarrow${\bf C03}$\rightarrow${\bf C01}$\rightarrow${\bf C04}$\rightarrow${\bf C04}$\rightarrow${\bf C02}$\rightarrow${\bf C01}$\rightarrow${\bf C03}$\rightarrow${\bf C01}$\rightarrow${\bf C02}$\rightarrow${\bf C01}$\rightarrow${\bf C05}$\rightarrow${\bf C05}$\rightarrow${\bf C03}$\rightarrow${\bf C01} \\ \hline
{\em Natta 5} & {\bf C07}$\rightarrow${\bf C07}$\rightarrow${\bf C06}$\rightarrow${\bf C06}$\rightarrow${\bf C02}$\rightarrow${\bf C01}$\rightarrow${\bf C02}$\rightarrow${\bf C01}$\rightarrow${\bf C07}$\rightarrow${\bf C07}$\rightarrow${\bf C06}$\rightarrow${\bf C06}$\rightarrow${\bf C03}$\rightarrow${\bf C01}$\rightarrow${\bf C03}$\rightarrow${\bf C01} \\ \hline
{\em Natta 6} & {\bf C08}$\rightarrow${\bf C08}$\rightarrow${\bf C10}$\rightarrow${\bf C10}$\rightarrow${\bf C02}$\rightarrow${\bf C01}$\rightarrow${\bf C02}$\rightarrow${\bf C01}$\rightarrow${\bf C09}$\rightarrow${\bf C09}$\rightarrow${\bf C11}$\rightarrow${\bf C11}$\rightarrow${\bf C03}$\rightarrow${\bf C01}$\rightarrow${\bf C03}$\rightarrow${\bf C01} \\ \hline
{\em Natta 7} & {\bf C12}$\rightarrow${\bf C12}$\rightarrow${\bf C14}$\rightarrow${\bf C14}$\rightarrow${\bf C07}$\rightarrow${\bf C07}$\rightarrow${\bf C06}$\rightarrow${\bf C06}$\rightarrow${\bf C13}$\rightarrow${\bf C13}$\rightarrow${\bf C15}$\rightarrow${\bf C15}$\rightarrow${\bf C07}$\rightarrow${\bf C07}$\rightarrow${\bf C06}$\rightarrow${\bf C06} \\ \hline
{\em Natta 8} & {\bf C16}$\rightarrow${\bf C16}$\rightarrow$\textcolor{red}{T01}$\rightarrow${\bf C18}$\rightarrow${\bf C19}$\rightarrow${\bf C21}$\rightarrow$\textcolor{red}{T02}$\rightarrow${\bf C23}$\rightarrow${\bf C17}$\rightarrow${\bf C17}$\rightarrow$\textcolor{red}{T03}$\rightarrow${\bf C18}$\rightarrow${\bf C20}$\rightarrow${\bf C22}$\rightarrow$\textcolor{red}{T04}$\rightarrow${\bf C23} \\ \hline

\multicolumn{2}{c}{ } \\ \hline
{\em Beats (full)} & \textcolor{red}{B01$\rightarrow$B02$\rightarrow$B03$\rightarrow$B04$\rightarrow$B05$\rightarrow$B06$\rightarrow$B07$\rightarrow$B08$\rightarrow$}\textcolor{blue}{B09$\rightarrow$B10$\rightarrow$B11$\rightarrow$B12$\rightarrow$B13$\rightarrow$B14$\rightarrow$B15$\rightarrow$B16$\rightarrow$} \\ 
            & \textcolor{red}{B17$\rightarrow$B18$\rightarrow$B19$\rightarrow$B20$\rightarrow$B21$\rightarrow$B22$\rightarrow$B23$\rightarrow$B24$\rightarrow$}\textcolor{blue}{B25$\rightarrow$B26$\rightarrow$B27$\rightarrow$B28$\rightarrow$B29$\rightarrow$B30$\rightarrow$B31$\rightarrow$B32} \\ \hline\hline
{\em Natta 4} & {\bf C02}$\rightarrow${\bf C01}$\rightarrow${\bf C02}$\rightarrow${\bf C01}$\rightarrow${\bf C03}$\rightarrow${\bf C01}$\rightarrow${\bf C03}$\rightarrow${\bf C01}$\rightarrow${\bf C04}$\rightarrow${\bf C04}$\rightarrow${\bf C04}$\rightarrow${\bf C04}$\rightarrow${\bf C02}$\rightarrow${\bf C01}$\rightarrow${\bf C02}$\rightarrow${\bf C01}$\rightarrow$\\ 
 & {\bf C03}$\rightarrow${\bf C01}$\rightarrow${\bf C03}$\rightarrow${\bf C01}$\rightarrow${\bf C02}$\rightarrow${\bf C01}$\rightarrow${\bf C02}$\rightarrow${\bf C01}$\rightarrow${\bf C05}$\rightarrow${\bf C05}$\rightarrow${\bf C05}$\rightarrow${\bf C05}$\rightarrow${\bf C03}$\rightarrow${\bf C01}$\rightarrow${\bf C03}$\rightarrow${\bf C01} \\ \hline
\multicolumn{2}{c}{ } \\
\multicolumn{2}{l}{$\bullet$ {\em Sollukattu} = {\em Natta}. Number of Beats in a bar = 8. {\em Bol}s with beats are:} \\
\multicolumn{2}{l}{\ \ \ \ B01 ({\em tai yum})$\rightarrow$B02 ({\em tat tat})$\rightarrow$B03 ({\em tai yum})$\rightarrow$B04 ({\em ta})$\rightarrow$B05 ({\em tai yum})$\rightarrow$B06 ({\em tat tat})$\rightarrow$B07 ({\em tai yum})$\rightarrow$B08 ({\em ta})} \\ 
\multicolumn{2}{l}{$\bullet$ Bars of the {\em Sollukattu} are shown in alternating colors} \\
\multicolumn{2}{l}{$\bullet$ A cycle of {\em Natta Adavu} has 16 or 32 postures and spans 2 or 4 bars of the {\em Sollukattu}} \\
\multicolumn{2}{l}{$\bullet$ {\em Natta 8 Adavu} has transition postures at beats B03, B07, B11, and B15. These are ignored here}
\end{tabular}
\end{footnotesize}
\end{table*}

\begin{table}[!ht]
\caption{Confusion matrix for {\em Adavu} (sequence) Recognition\label{tab:seq}}
\centering 
\begin{tabular}{cc|cccccccc|c|} 
&\multicolumn{1}{c}{} & \multicolumn{8}{c}{\bf Predicted Class} \\
&\multicolumn{1}{c}{} & 1 & 2 & 3 & 4 & 4 & 6 & 7 & \multicolumn{1}{c}{8} & \multicolumn{1}{c}{Total}  \\ \cline{3-11}
\multirow{14}{*}{\textbf{\begin{rotate}{+90}Actual Class\end{rotate}}} 
& 1 & 4 & \alert{\bf 3} & 0 & 0 & 0 & 0 & 0 & 0 & 7 \\
& 2 & 0 & 7 & 0 & 0 & 0 & 0 & 0 & 0 & 7 \\
& 3 & 0 & 0 & 7 & 0 & 0 & 0 & 0 & 0 & 7 \\
& 4 & 0 & 0 & 0 & 7 & 0 & 0 & 0 & 0 & 7 \\
& 5 & 0 & 0 & 0 & 0 & 7 & 0 & 0 & 0 & 7 \\
& 6 & 0 & 0 & 0 & 0 & 0 & 7 & 0 & 0 & 7 \\
& 7 & 0 & 0 & 0 & 0 & 0 & 0 & 7 & 0 & 7 \\
& 8 & 0 & 0 & 0 & 0 & 0 & 0 & 0 & 7 & 7 \\ \cline{3-10}
&   & 4 & 10 & 7 & 7 & 7 & 7 & 7 & 7 & \\ \cline{3-11}
\end{tabular}
\end{table}


\section{Conclusions\label{sec:conclusion}}
In this paper, we have addressed a few fundamental problems of dance analysis related to dance video segmentation and key posture extraction, key posture recognition and dance sequence recognition.
Dance video segmentation is often subjective and critically depends on the style of dance. Here we have focused on {\em Bharatanatyam} and  created a model for analyzing {\em Adavus}, an essential unit of {\em Bharatanatyam}, with the help of  accompanying  audio, video and sync events
We use the audio events to first extract {\em Key Frame}s. The {\em Key Posture}s are then identified from the extracted K-frames. Finally, from the sequence of postures we recognize the {\em Adavu}. Posture recognition problem is addressed by using three types of classifiers -- namely GMM, SVM, and CNN. Although these classifiers use different sets of features and have their respective strengths and limitations, overall more than 90\% accuracy is achieved. We achieve 83.5\% accuracy in extracting {\em K-frame}s. For posture recognition we identify 23 posture classes that occur in {\em Natta Adavu}s. We achieve 83\% accuracy for posture recognition using GMM with angular features from skeletons and 98\% accuracy  using SVM with HOG features from RGB frames. Since some of posture classes are very similar or have very little data, we merge 23 classes into 15 for building a CNN classifier. We achieve an accuracy of 99\% for posture recognition using CNN.


We have done the structural analysis of dance considering the underlying semantics. Hence, the algorithms developed in this paper can be used in many applications including:

\begin{itemize}
\item {\bf Dance Transcription}: Forms like {\em Bharatanatyam} in ICD is traditionally passed on from experts to their disciples through personal contact and face-to-face training. Consequently, preservation of the heritage of ICD is a major challenge as it is often threatened with the possibility of this trainer-trainee human-chain breaking down. The need of the hour, therefore, is to transcribe ICD in a notation (much in the way music in transcribed in a notation) that can transcend time. {\em Labanotation}, commonly used in the Western dance forms to graphically depict dance steps.
{\em Bharatanatyam} {\em Adavu}s are transcribed to {\em Labanotation} in~\cite{Sankhla2018} and represented in a parseable XML representation of the graphical notation. 

\item {\bf Dance Tutoring}: A dance trainee, as a user, will be able to use the tutoring system to check the correctness of her {\em Adavu} performances against pre-recorded performances by the experts. We have built a rudimentary version of this application~\cite{Aich2018} using Dynamic Time Warping (DTW) to associate posture frames from expert's performance video with the corresponding ones from the trainee's. This should be developed as a complete application to improve training.

\item {\bf Dance Animation}: Once the sequencing information has been extracted (a subset of the {\em Adavu} recognition problem), ICD performances can be synthetically created through animation of {\em avatar}s as well as actual 3D human models.
\end{itemize}

\section*{Acknowledgment}
The work of the first author is supported by TCS Research Scholar Program of Tata Consultancy Services of India.

{\small
\bibliographystyle{plain}
\bibliography{References}
}

\end{document}